\documentclass{article}
\usepackage{iclr2027_conference,times}

\usepackage[utf8]{inputenc}
\usepackage[T1]{fontenc}
\usepackage{hyperref}
\usepackage{url}
\usepackage{microtype}
\usepackage{xcolor}
\usepackage[normalem]{ulem}  
\definecolor{revblue}{RGB}{0,86,190}

\usepackage{amsmath}
\usepackage{amsfonts}
\usepackage{amssymb}
\usepackage{nicefrac}

\usepackage{booktabs}
\usepackage{graphicx}
\usepackage{makecell}
\usepackage{multirow}
\usepackage{wrapfig}
\usepackage{pifont}          
\usepackage{caption}
\usepackage{subcaption}
\usepackage[section]{placeins}

\title{Fixing the Last Mile: Failure-Aware Image Restoration with
Region-Specific Guidance}\date{}

\author{%
  Hongyu An$^{1}$,
  Fanghua Yu$^{2}$,
  Kainan Yan$^{3}$,
  Luc Van Gool$^{1}$, Chao Dong$^{3,4}$,
  Jinjin Gu$^{1}$ \\[0.5em]
  \normalfont
  \footnotesize
  $^{1}$~INSAIT, Sofia University ``St. Kliment Ohridski'' \\
  \footnotesize
  $^{2}$~The University of Hong Kong \\
  \footnotesize
  $^{3}$~Shenzhen Institutes of Advanced Technology, Chinese Academy of Sciences \\ \footnotesize $^{4}$~Shenzhen University of Advanced Technology
}
\iclrfinalcopy

\begin{document}

\maketitle
\lhead{Preprint}  

\begin{abstract}
Recent advances in image restoration have greatly improved global image fidelity, producing visually plausible results across most regions of an image.
Nevertheless, even high-quality restorations often exhibit localized failures, including artifacts in semantically important regions such as faces and text, structural inconsistencies, and texture mismatches.
Although these errors typically affect only a small portion of the image, they often dominate the overall perceptual quality of the result.
In this work, we propose \textbf{Failure-Aware Image Restoration (FAIR)} to address the \textbf{``last-mile''} problem in image restoration.
Rather than applying one uniform generative prior over the whole image, we condition a single-pass restoration model on both \textbf{where} local failures are likely to occur and \textbf{what} corrective guidance each region requires.
This region-bound conditioning enables fine-grained and controllable restoration that improves semantic accuracy and stylistic consistency in critical regions while maintaining faithful, high-quality restoration elsewhere.
Our method complements existing restoration pipelines by shifting the emphasis from global fidelity to local perceptual acceptability.
Extensive experiments show that the proposed approach resolves common failure cases and improves perceptual quality, bridging the gap between technical accuracy and practical usability.
\end{abstract}

\section{Introduction}
\label{sec:intro}
Recent advances in perception-oriented image restoration~\citep{lin2024diffbir, yu2024supir, wu2024seesr, wu2024osediff, chen2025faithdiff, lin2025hypir} have substantially improved overall image quality, producing results that are globally plausible and perceptually rich.
%
%
While existing methods have demonstrated impressive performance, we identify an overlooked class of \textbf{``last-mile''} problems in image restoration that fundamentally hinders their real-world deployment.

\begin{figure}[!t]
    \centering
    \includegraphics[width=1.0\linewidth]{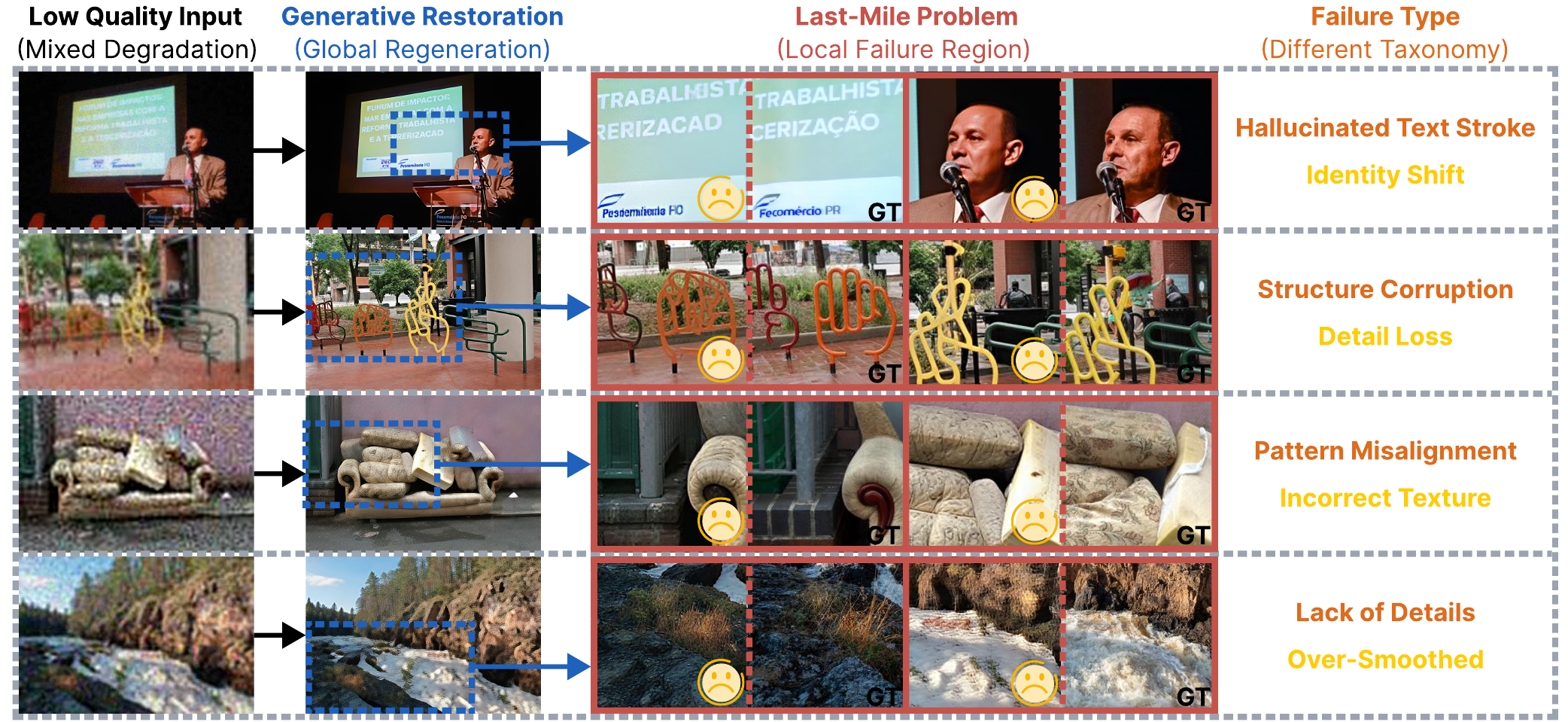}
    \vspace{-8pt}
    \caption{\textbf{The ``last-mile'' problem in generative image restoration.} State-of-the-art methods~\citep{yu2024supir, wu2024seesr, lin2025hypir} produce globally plausible reconstructions from severely degraded inputs, yet localized failures persist. Despite their small spatial extent, these failures dominate perceived quality and practical acceptability.}
    \vspace{-12pt}
\label{fig:failure_cases}
\end{figure}

A near-perfect landscape with a distorted face, or a sharp document with garbled characters, is rarely considered acceptable regardless of how well the remaining regions are recovered.
As shown in Fig.~\ref{fig:failure_cases}, such localized failures arise consistently under severe degradations, where heavy blur, noise, or compression remove local information beyond recovery, forcing generative priors to hallucinate plausible yet unfaithful content.
Although these errors typically affect only a small portion of the image, they often dominate the perceived quality of the result.
We refer to this challenge as the \textbf{``last-mile''} problem in image restoration: closing the perceptual gap by predicting failure-prone regions from the low-quality input and guiding their restoration with external references.

%
%

The last-mile problem stems from a structural limitation of current restoration pipelines.
Resolving it requires two capabilities that existing methods lack: external information to recover content that is irretrievable from the degraded input, and fine-grained local control to apply such information only where it is needed.
Yet existing approaches typically cast restoration as a global reconstruction problem, without explicitly modeling \textbf{where} local failures are likely to occur and, within each at-risk region, \textbf{what} type of failure it is prone to and what corrective guidance it requires.
Consequently, the same generative prior is applied uniformly across the image, from smooth backgrounds to semantically important areas such as faces, text, and logos.

The challenge is to introduce external information selectively: it should help regions with insufficient recoverable content without unnecessarily altering regions well constrained by the input.
Even recent semantic-guided methods~\citep{yu2024supir, wu2024seesr, sun2024coser} introduce high-level priors at the image level, while deriving them largely from the corrupted input.
When local information is severely lost, these priors can generate plausible details without ensuring fidelity to the original content.
Restoration therefore needs evidence beyond the input and spatial control to apply it only where needed.
Deferring guidance to post-hoc repair has further costs: it may disturb well-restored regions, while the second model must work from an image already altered by the first, potentially missing original cues or containing new artifacts.

\begin{wrapfigure}{r}{0.68\linewidth}
    \centering
    \vspace{-8pt}
    \includegraphics[width=\linewidth]{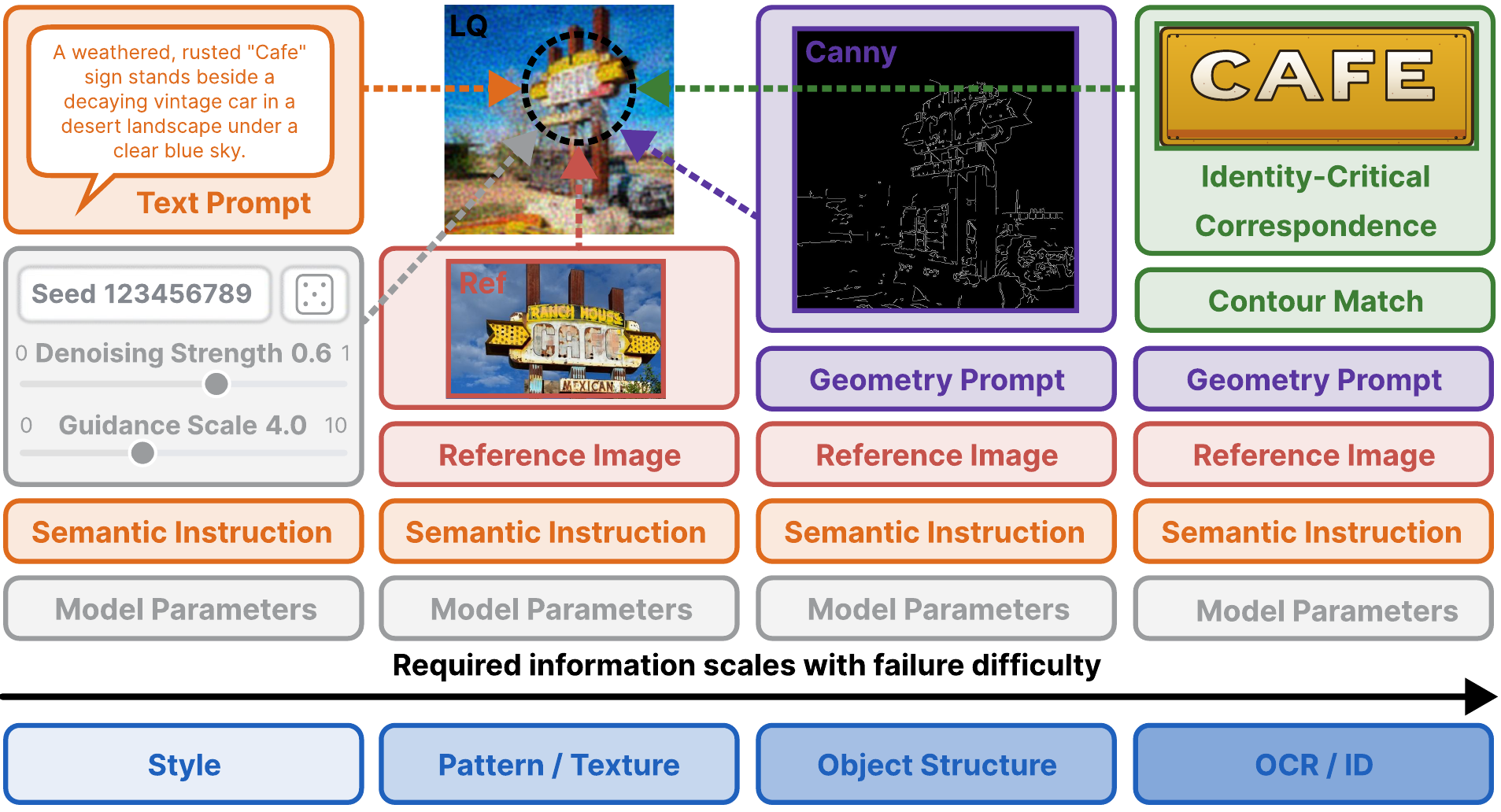}
    \vspace{-8pt}
    \caption{\textbf{Failure-aware guidance for local restoration challenges.} We group typical local restoration failures into four categories (bottom row) and show the corresponding condition combinations (top). As local content becomes harder to infer from the low-quality input, guidance progresses from semantic instructions and parameter controls to reference images, geometric contours, and identity correspondences. Region control limits where these conditions act.}
    \vspace{-5pt}
    \label{fig:valuable_approach}
\end{wrapfigure}

We propose \textbf{FAIR}, a failure-aware image restoration framework that analyzes the low-quality input, predicts failure-prone regions and their types, and prepares matching conditions before restoration.
Rather than treating the image as a uniform field, FAIR recognizes that under severe degradation, small but perceptually critical regions often require different forms of corrective information.
To study this behavior systematically, we construct a large-scale failure-aware dataset by applying state-of-the-art restoration models to degraded inputs and automatically annotating both \textbf{where} failures occur and \textbf{what} type of failure is present.
Each failure region is additionally paired with an appropriate problem description, yielding region-level supervision aligned with our failure taxonomy.
We observe a consistent pattern in this dataset: restoration failures cluster into a small set of semantically meaningful categories, each requiring different forms and strengths of external guidance.

As illustrated in Fig.~\ref{fig:valuable_approach}, we organize restoration guidance by increasing constraint strength, with more specific guidance addressing progressively harder local failures: (i) semantic instructions and parametric controls for subjective artifacts; (ii) reference exemplars for pattern and texture inconsistencies; (iii) geometric prompts for fine structural correction; and (iv) identity correspondences for identity-critical regions.
Different failures call for different forms of guidance, and FAIR accepts each of them as a condition.
Unlike existing methods that apply uniform generation across the image, FAIR binds each guidance condition to its target region.
Built on FLUX.2-klein-9B~\citep{flux2_2025} with lightweight LoRA modules, FAIR uses region control to specify \textbf{where} additional guidance is provided and a semantic reference to indicate \textbf{what} the restoration should follow.
Training adopts a two-stage curriculum: single-step restoration pretraining, followed by region-anchored fine-tuning with external guidance, spatial control, and region-aware reconstruction reweighting.
Experiments show that FAIR improves perceptual acceptability over strong baselines while preserving quality~elsewhere.


\begin{figure}[!t]
    \centering
    \includegraphics[width=1.0\linewidth]{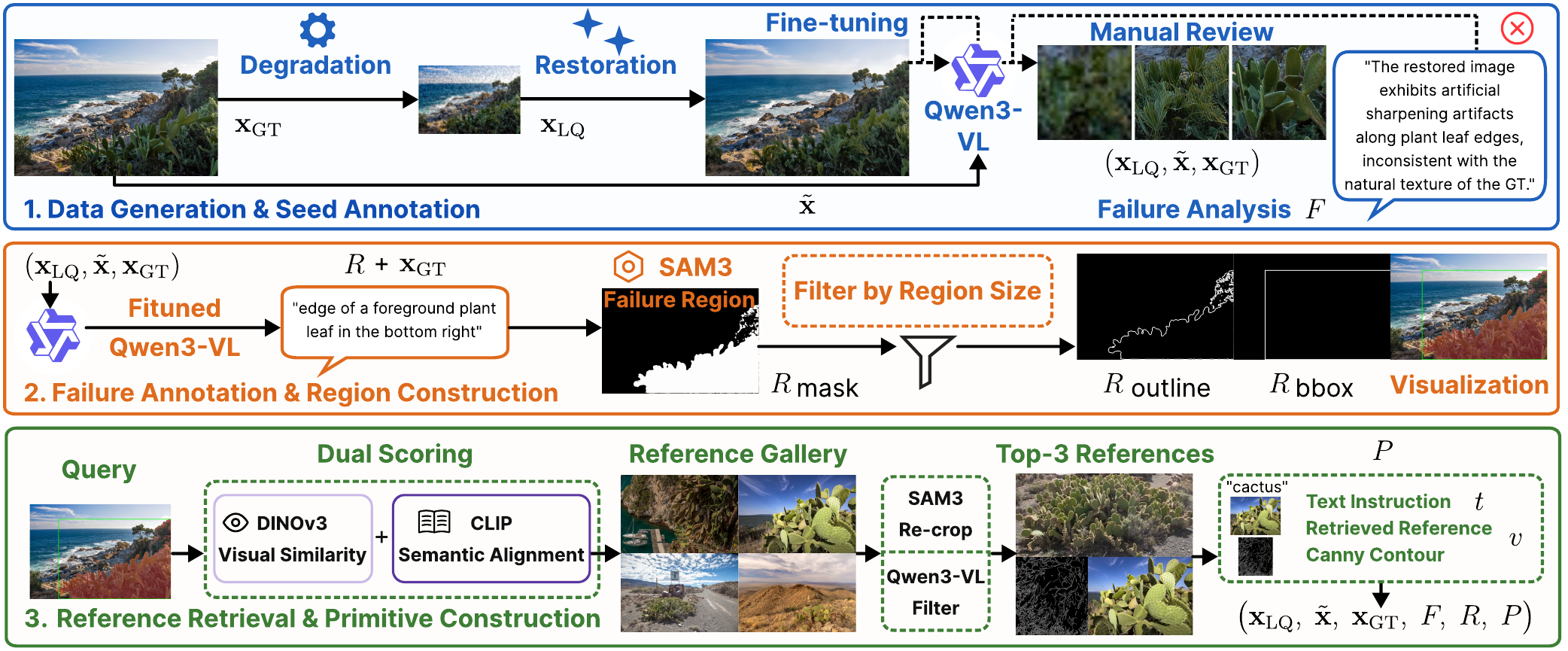}
    \vspace{-10pt}
    \caption{\textbf{Three-stage offline construction of failure-aware training data.} Stage I degrades $\mathbf{x}_{\mathrm{GT}}$ to $\mathbf{x}_{\mathrm{LQ}}$, produces a coarse restoration $\tilde{\mathbf{x}}$, and uses screened annotations to fine-tune Qwen3-VL-8B. Stage II supplies $(\mathbf{x}_{\mathrm{LQ}},\tilde{\mathbf{x}},\mathbf{x}_{\mathrm{GT}})$ to the fine-tuned model to identify failure type $F$ and a localization phrase; SAM3 converts the phrase into region control $R$. Stage III retrieves and filters local references to build restoration guidance $P=(t,v)$, with text instruction $t$ and optional visual guidance $v$.}
    \vspace{-8pt}
    \label{fig:pipeline}
\end{figure}\vspace{-6pt}
\section{Related Work}
\vspace{-4pt}

\noindent\textbf{Image Restoration.}\quad
Recent progress in image restoration has been largely driven by deep learning, leading to substantial improvements in super-resolution, denoising, and deblurring.
Early CNN-based approaches~\citep{dong2015srcnn, lim2017edsr, zhang2018rcan, 
dai2019san, mei2021image} learn direct mappings from degraded inputs to clean images and achieve strong performance under synthetic degradations.
Transformer-based methods~\citep{liang2021swinir, wang2022uformer, 
zamir2022restormer, chen2023hat, wu2025dswinir} further improve reconstruction quality through long-range dependency modeling and more expressive architectures.
GAN-based approaches~\citep{ledig2017srgan, 
wang2018esrgan, wang2021realesrgan, liang2022details} focus on perceptual realism by synthesizing high-frequency details through adversarial training.
Despite these advances, most existing methods remain optimized for global reconstruction objectives such as pixel-level fidelity or perceptual similarity.
As a result, they often fail in localized regions, particularly semantically important areas such as faces and text, where even minor artifacts can substantially degrade perceptual quality.

\noindent\textbf{Diffusion-Based Restoration.}\quad
Diffusion models~\citep{ho2020ddpm, song2020ddim, dhariwal2021diffusion, 
rombach2022ldm} have become a dominant paradigm for image restoration~\citep{lin2024diffbir, yu2024supir, wu2024osediff, 
chen2025faithdiff, lin2025hypir, saharia2022sr3, dong2025tsd}, achieving strong perceptual quality across a wide range of tasks.
However, image-level conditioning alone often struggles under severe degradations.
Recent studies report a shift from under-generation to over-generation, where strong generative priors produce hallucinations, structural distortions, and semantic inconsistencies in challenging regions such as faces, fine structures, and textured surfaces~\citep{yin2026far}.
To address these issues, several methods introduce explicit semantic guidance.
PASD~\citep{yang2024pasd} extracts semantic cues from low-resolution inputs using off-the-shelf vision models and injects them through pixel-aware cross-attention.
SeeSR~\citep{wu2024seesr} generates degradation-aware tag prompts, while SUPIR~\citep{yu2024supir} and CoSeR~\citep{sun2024coser} leverage multimodal LLMs to produce richer semantic descriptions.
InstructRestore~\citep{liu2025instructrestore} further introduces user-instruction control, and CODSR~\citep{chen2025codsr} reformulates restoration as one-step generation with region-adaptive priors.
Nevertheless, these approaches remain primarily driven by global objectives and image-level priors, without explicitly predicting \textbf{where} restoration is likely to fail or what guidance each region needs.

\noindent\textbf{Reference-Based Restoration.}\quad
Reference-based restoration uses external high-quality images to recover details irretrievable from severely degraded inputs.
Early methods rely on patch matching and texture transfer~\citep{zhang2019srntt, yang2020ttsr, lu2021masa, jiang2021wrsr, cao2022datsr}, typically assuming curated, strongly aligned references.
Recent diffusion-based approaches relax this requirement: ReFIR~\citep{guo2024refir} injects retrieved features through cross-image attention, RASR~\citep{yan2025rasr} formulates retrieval-augmented reference super-resolution, and AdaRefSR~\citep{wang2026adarefsr} adaptively weights reference features according to local relevance.
Nevertheless, these methods use references as a primary condition for overall reconstruction without explicitly modeling failure-prone regions.
FAIR instead binds guidance to the relevant failure-prone region, confining external evidence there.

\section{Method}
\label{sec:method}

Given a low-quality input $\mathbf{x}_{\mathrm{LQ}}$, we produce a restored image $\hat{\mathbf{x}}$ by applying dedicated guidance to failure-prone regions while restoring the remaining regions without additional guidance:
\begin{equation}
\hat{\mathbf{x}}
=
f_{\theta}\bigl(
\mathbf{x}_{\mathrm{LQ}},R,P
\bigr).
\label{eq:task}
\end{equation}
Here, $R$ specifies \textbf{where} to provide additional guidance using a localization phrase,
bounding box, outline, or mask. The restoration guidance $P=(t,v)$ specifies \textbf{what} information to provide, combining a textual instruction $t$ with optional visual guidance $v$ from a retrieved reference or Canny contour; Appendix~\ref{app:dataset_details} maps the guidance types of Fig.~\ref{fig:valuable_approach} onto $P$.
At inference, $R$ and $P$ are predicted from
$\mathbf{x}_{\mathrm{LQ}}$ or provided by a user.
Let $\mathcal{F}$ denote the four failure tiers of
Fig.~\ref{fig:valuable_approach}; $F\in\mathcal{F}$ indicates which form of $P$ a region typically needs. Binding $P$ to $R$ provides additional information to failure-prone regions during restoration.
FAIR comprises a failure-aware dataset
(Sec.~\ref{sec:data}), localized multimodal conditioning
(Sec.~\ref{sec:arch}), and a two-stage region-aware
curriculum (Sec.~\ref{sec:curriculum}).

\subsection{Fine-Grained Failure-Aware Data Curation}
\label{sec:data}

\begin{wrapfigure}[18]{r}{0.4\textwidth}
    \vspace{-14pt}
    \centering
    \includegraphics[width=0.4\textwidth]{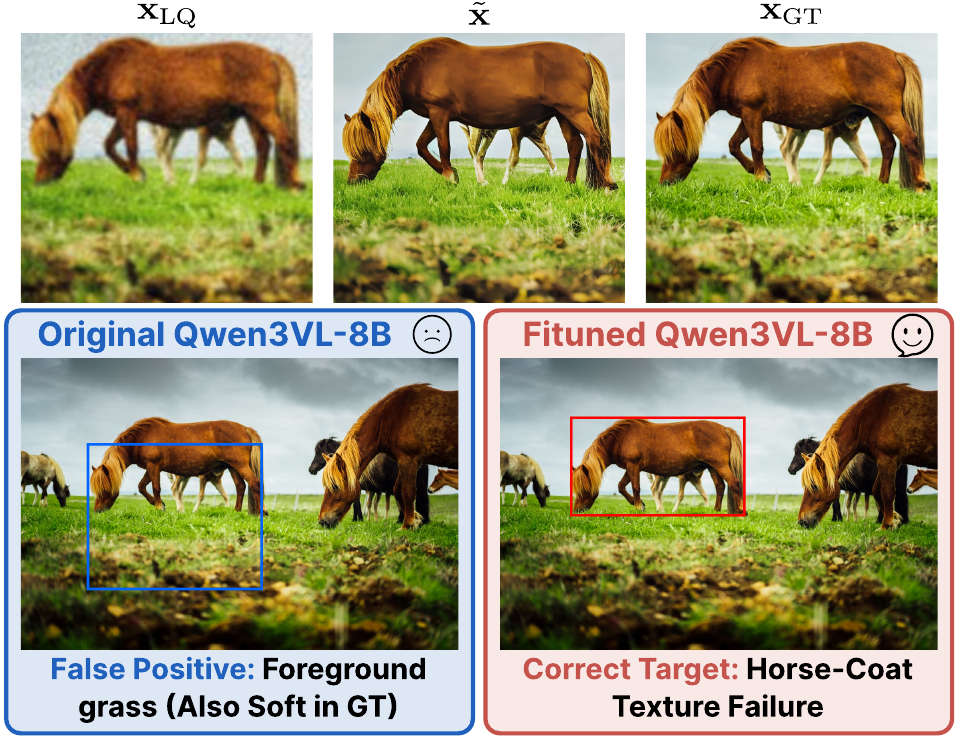}
    \vspace{-14pt}
    \caption{\textbf{Failure localization before and after fine-tuning.} The off-the-shelf model selects foreground grass (blue), while the fine-tuned model selects the central horse's mane and tail (red).}
    \label{fig:qwen_finetune}
    \vspace{-6pt}
\end{wrapfigure}

Training a failure-aware model requires supervision beyond standard
$(\mathbf{x}_{\mathrm{LQ}},\mathbf{x}_{\mathrm{GT}})$ pairs. For each pair,
we generate a coarse restoration $\tilde{\mathbf{x}}$ and structured
annotations $(F,R,P)$ specifying the failure type, target region, and
corresponding restoration guidance:
\begin{equation}
\bigl(\mathbf{x}_{\mathrm{LQ}},\;
\tilde{\mathbf{x}},\;\mathbf{x}_{\mathrm{GT}},\;F,\;R,\;P\bigr).
\end{equation} As illustrated in Fig.~\ref{fig:pipeline}, our failure-aware data construction pipeline proceeds in three stages.

\noindent\textbf{Stage I: Data generation and seed annotation.}
We draw high-quality images $\mathbf{x}_{\mathrm{GT}}$ from LSDIR and
Unsplash~\citep{li2023lsdir,unsplash}, with FFHQ~\citep{karras2019ffhq}
added to the restoration mix for face coverage, synthesize
$\mathbf{x}_{\mathrm{LQ}}$ using the second-order Real-ESRGAN
pipeline~\citep{wang2021realesrgan}, and process it with multiple
state-of-the-art restoration models to obtain coarse restorations
$\tilde{\mathbf{x}}$ exhibiting realistic local failures.
For each $(\mathbf{x}_{\mathrm{LQ}},\tilde{\mathbf{x}},
\mathbf{x}_{\mathrm{GT}})$ triplet, Qwen3-VL-8B~\citep{bai2025qwen3}
predicts a failure label $F$ and a phrase identifying the affected region.
A structured prompt defines the failure taxonomy, output format, and
localization requirements.
Because the off-the-shelf model may mislocalize failures
(Fig.~\ref{fig:qwen_finetune}), we manually verify its outputs against
the ground truth and retain 1K triplet--annotation pairs with correct
failure labels and localization. These verified pairs are used to
fine-tune Qwen3-VL-8B with rank-$64$ LoRA, yielding the annotator used
in Stage~II.

\noindent\textbf{Stage II: Failure annotation and region construction.}
The fine-tuned annotator processes the full pool with the same input triplet,
producing a failure label and localization phrase. \mbox{SAM3~\citep{carion2025sam3}}
segments the corresponding region in $\mathbf{x}_{\mathrm{GT}}$; after
filtering masks with extreme area ratios, we derive a bounding box and closed
outline. The phrase, box, outline, and mask form four interchangeable controls
$R$; after the Stage-III filtering of Appendix~\ref{app:reference_retrieval},
$67$K regions remain (Appendix~\ref{app:stats}). At test time, condition
prediction uses $\mathbf{x}_{\mathrm{LQ}}$ alone; its accuracy and
GT-assisted oracle are evaluated in Appendix~\ref{app:robustness}.

\begin{figure}[!t]
    \centering
    \includegraphics[width=1.0\linewidth]{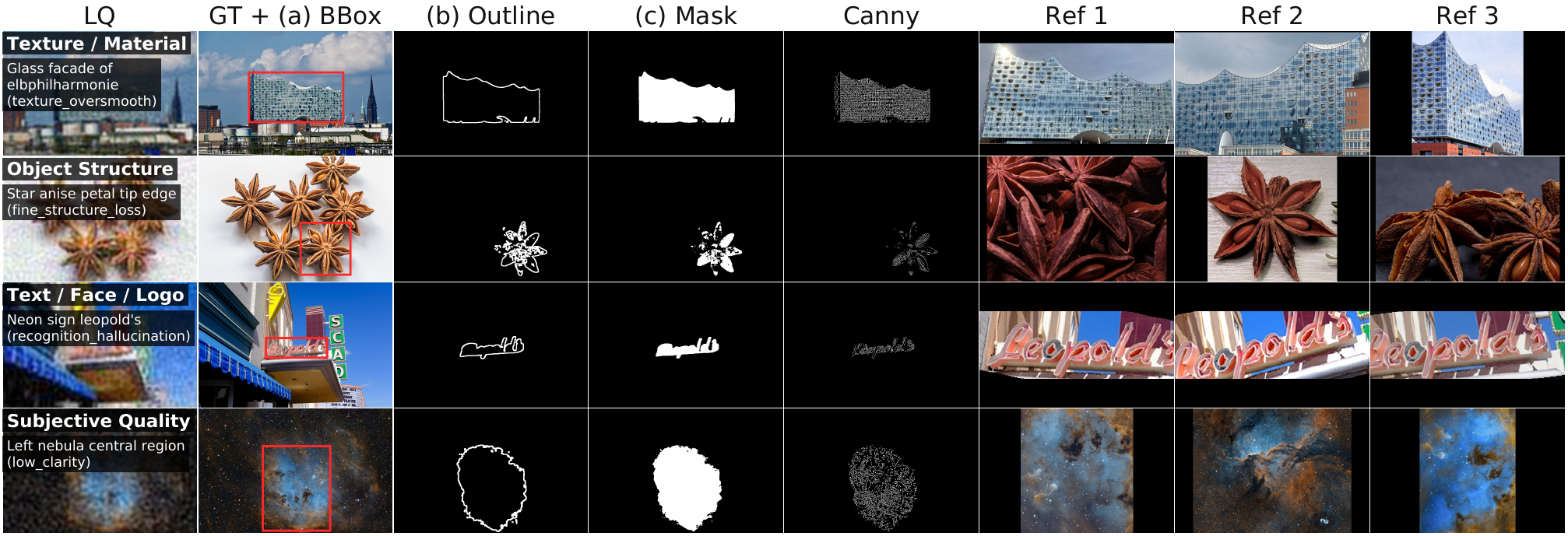}
    \vspace{-10pt}
    \caption{\textbf{Representative failure-aware dataset samples, one per tier.} Each shows an LQ--GT pair, four region controls (localization phrase, box, outline, mask), restoration guidance $P=(t,v)$ (textual instruction $t$ with optional visual guidance $v$), and failure label $F$.}
    \vspace{-10pt}
    \label{fig:dataset}
\end{figure}

\noindent\textbf{Stage III: Reference retrieval and guidance construction.}
For each annotated region, we retrieve candidate patches from a multi-scale
LSDIR--Unsplash gallery, ranked by DINOv3~\citep{simeoni2025dinov3} similarity
to the target region and CLIP~\citep{radford2021clip} consistency with its
semantic description. SAM3 re-crops the corresponding instances, while
Qwen3-VL-8B filters references inconsistent with the region description.
We retain up to three cross-image references from distinct source images. The restoration guidance $P=(t,v)$ combines textual instruction
$t$ with visual guidance $v$ sampled as empty, a visual reference, or a Canny
contour. Appendix~\ref{app:reference_retrieval} details the retrieval,
filtering, and guidance-selection procedures.

\noindent\textbf{Dataset summary.}
The resulting dataset contains 67K annotated failure regions spanning
the four failure tiers. Each region is associated with a failure label
$F$ and multiple compatible forms of region control $R$ and restoration
guidance $P$. During training, one compatible $(R,P)$ combination is
randomly sampled for each region, exposing the model to diverse spatial
and corrective conditions.
Fig.~\ref{fig:dataset} shows representative samples from each tier;
additional examples, category distributions, and dataset statistics
are provided in Appendix~\ref{app:dataset_details}.

\subsection{Two-Stage Region-Guided Restoration Framework}
\label{sec:framework}

FAIR adapts FLUX.2-klein-9B~\citep{flux2_2025} for restoration
conditioned on the low-quality image, region control $R$, and restoration
guidance $P$:
\[
\hat{\mathbf{x}} \sim p_\theta\!\bigl(\hat{\mathbf{x}} \,\big|\,
\mathbf{x}_{\mathrm{LQ}},R,P\bigr)
\]
using a fixed-timestep flow-matching formulation~\citep{lipman2022flow,esser2024scaling}. The guidance $P=(t,v)$ is defined in Sec.~\ref{sec:data}, and $R$ localizes its effect.

\subsubsection{Architecture and localized guidance conditioning}
\label{sec:arch}

\begin{figure}[!t]
    \centering
    \includegraphics[width=0.95\linewidth]{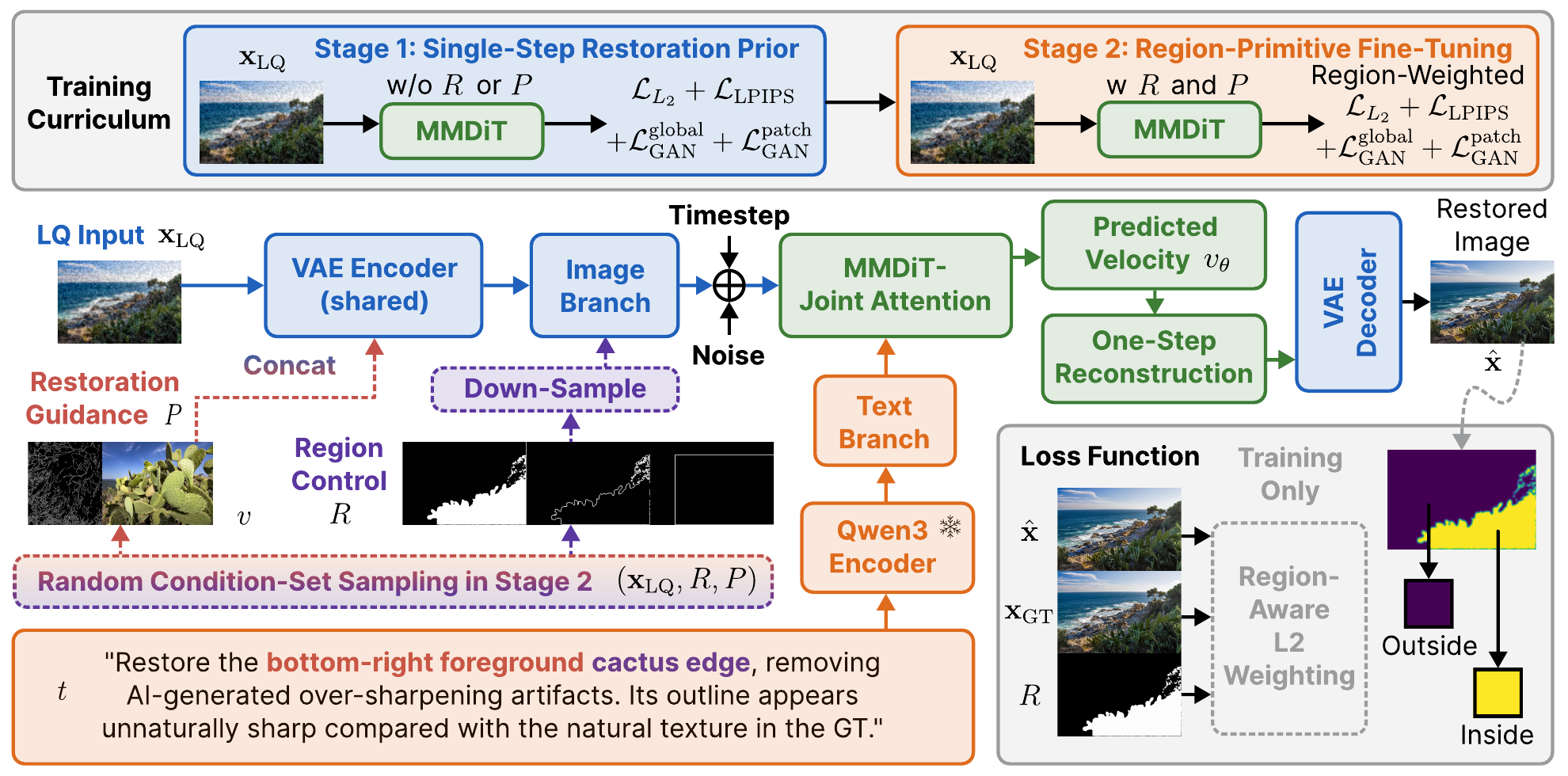}
    \vspace{-2pt}
    \caption{\textbf{Overview of FAIR's two-stage training.} Stage~1 learns a general restoration prior. Stage~2 conditions restoration on region control $R$ and guidance $P=(t,v)$, using region-aware supervision to bind the guidance to its target region.}
    \vspace{-8pt}
    \label{fig:framework}
\end{figure}

As illustrated in Fig.~\ref{fig:framework}, FAIR jointly conditions restoration on the low-quality input $\mathbf{x}_{\mathrm{LQ}}$, region control $R$, and restoration guidance $P$: the instruction $t$ enters the text stream, while spatial forms of $R$ and visual guidance $v$ enter the image stream, so that $R$ localizes the guidance (Appendix~\ref{app:stage2}).
The region map and $v$ are VAE-encoded and appended to the $\mathbf{x}_{\mathrm{LQ}}$ tokens as separate reference images with their own positional offsets, so that every condition reaches the target through the backbone's joint self-attention over a single token stream.

\subsubsection{Two-stage curriculum}
\label{sec:curriculum}

\noindent\textbf{Stage 1: general restoration prior.}
We fine-tune the LoRA-adapted backbone on LQ--GT pairs at $1024^2$ resolution. At a fixed timestep, the model predicts $v_\theta$ from noisy LQ latent $z_t$, reconstructs $z_0=z_t-\sigma_t v_\theta$, and decodes it as $\hat{\mathbf{x}}=\mathcal{D}(z_0)$. The generator objective is
\begin{equation}
\mathcal{L}^{\mathrm{S1}}_{G}
= \lambda_{L_2}\mathcal{L}_{L_2}
+ \lambda_{\mathrm{LPIPS}}\mathcal{L}_{\mathrm{LPIPS}}
+ \lambda_{\mathrm{GAN}}\mathcal{L}_{\mathrm{GAN}},
\label{eq:stage1}
\end{equation}
All three losses act on the decoded image. The GAN loss uses an $\mathrm{R1}$-regularized ConvNeXt discriminator. Loss weights and regularization settings are given in Appendix~\ref{app:training_details}.

\noindent\textbf{Stage 2: region-guided fine-tuning.}
We fine-tune the shared LoRA parameters at $512^2$ to condition restoration on region control $R$ and guidance $P=(t,v)$ (Sec.~\ref{sec:arch}), using the region-aware objective (Eq.~\eqref{eq:rec}). Each batch uses a $4{:}1$ mix of failure-aware and standard restoration samples. The failure-aware branch samples a localization phrase, bounding box, outline, or mask for $R$, with prompt-only dropout, and samples $v$ from \{empty, retrieved reference, Canny contour\}. Standard restoration samples retain the stage~1 prior (Appendix~\ref{app:training_details}).

\subsubsection{Region-aware reconstruction objective}
\label{sec:loss}

Uniform reconstruction weighting limits flexibility in target regions,
while weak supervision outside them can reduce fidelity. We weight the
target less than its surroundings by converting $R$ to a binary
mask, Gaussian-smoothing it into $m:\Omega\to[0,1]$, and reweighting only
the image-space $L_2$ term:
\begin{equation}
\begin{aligned}
w(\mathbf{u})
&=1+\alpha_{\mathrm{in}}m(\mathbf{u})
+\alpha_{\mathrm{out}}\bigl(1-m(\mathbf{u})\bigr),\\
\mathcal{L}^{\mathrm{region}}_{L_2}(\theta)
&=\mathbb{E}\!\left[
\mathop{\mathrm{mean}}_{\mathbf{u}\in\Omega}
w(\mathbf{u})
\left\|\hat{\mathbf{x}}(\mathbf{u})
-\mathbf{x}_{\mathrm{GT}}(\mathbf{u})\right\|_2^2
\right].
\end{aligned}
\label{eq:rec}
\end{equation}

The target region retains its standard reconstruction weight, while surrounding content receives a higher weight; Gaussian smoothing ensures a gradual transition. LPIPS and the adversarial loss remain image-wide, and $w=1$ everywhere without localization. See Appendix~\ref{app:stage2} for settings.



\section{Experiments}
\label{sec:experiments}

\noindent\textbf{Test sets.} We evaluate on three test sets: (i) $120$ synthetic triplets $(\mathbf{x}_{\text{LQ}},\mathbf{x}_{\text{GT}},v)$ constructed with the pipeline in Sec.~\ref{sec:data}, $30$ per tier, using second-order Real-ESRGAN degradation~\citep{wang2021realesrgan} and $v$ from semantically related samples; (ii) the paired WR-SR~\citep{jiang2021wrsr} benchmark with LQ--GT images and curated references; and (iii) an in-the-wild set without GT~\citep{yin2026far}. Conditions are obtained automatically (Appendix~\ref{app:benchmark}). On the synthetic set, Table~\ref{tab:quant_comparison_synthetic} simulates user-supplied guidance: $R$ is predicted from $\mathbf{x}_{\mathrm{LQ}}$, $t$ is the annotated instruction, and $v$ is retrieved with the clean GT crop of the region. We also evaluate fully LQ-driven guidance and generic restoration. Inference caps the long image side at 1024\,px.

\noindent\textbf{Baselines.} We compare five methods that do not use $v$: SeeSR~\citep{wu2024seesr}, SUPIR~\citep{yu2024supir}, HYPIR~\citep{lin2025hypir}, InstructRestore~\citep{liu2025instructrestore} (given the same region), and CODSR~\citep{chen2025codsr}. We further compare three reference-based methods given the same $v$: ReFIR+SeeSR, ReFIR+SUPIR~\citep{guo2024refir}, and AdaRefSR~\citep{wang2026adarefsr}.

\noindent\textbf{Metrics.} We report 13 metrics (Table~\ref{tab:quant_comparison_synthetic}): PSNR, SSIM~\citep{wang2004ssim}, FID~\citep{heusel2017fid}, DISTS~\citep{ding2020dists}, MUSIQ~\citep{ke2021musiq}, CLIP-IQA~\citep{wang2023clipiqa}, Q-Align~\citep{wu2023qalign}, MANIQA~\citep{yang2022maniqa}, TOPIQ~\citep{chen2024topiq}, DINO- and CLIP-Sim-Ref against $v$, and LPIPS\textsubscript{out}/LPIPS\textsubscript{in}~\citep{zhang2018lpips} outside/inside the GT mask.

\begin{figure*}[t]
  \centering
  \vspace{-4pt}
  \includegraphics[width=0.87\linewidth]{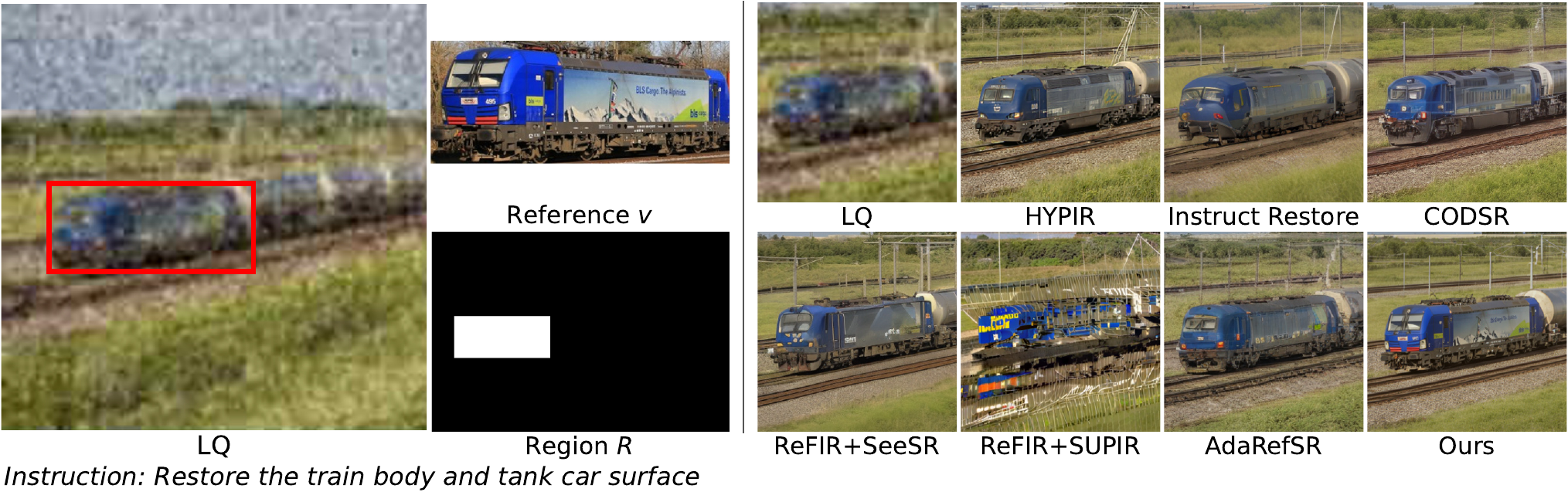}\\
  \includegraphics[width=0.87\linewidth]{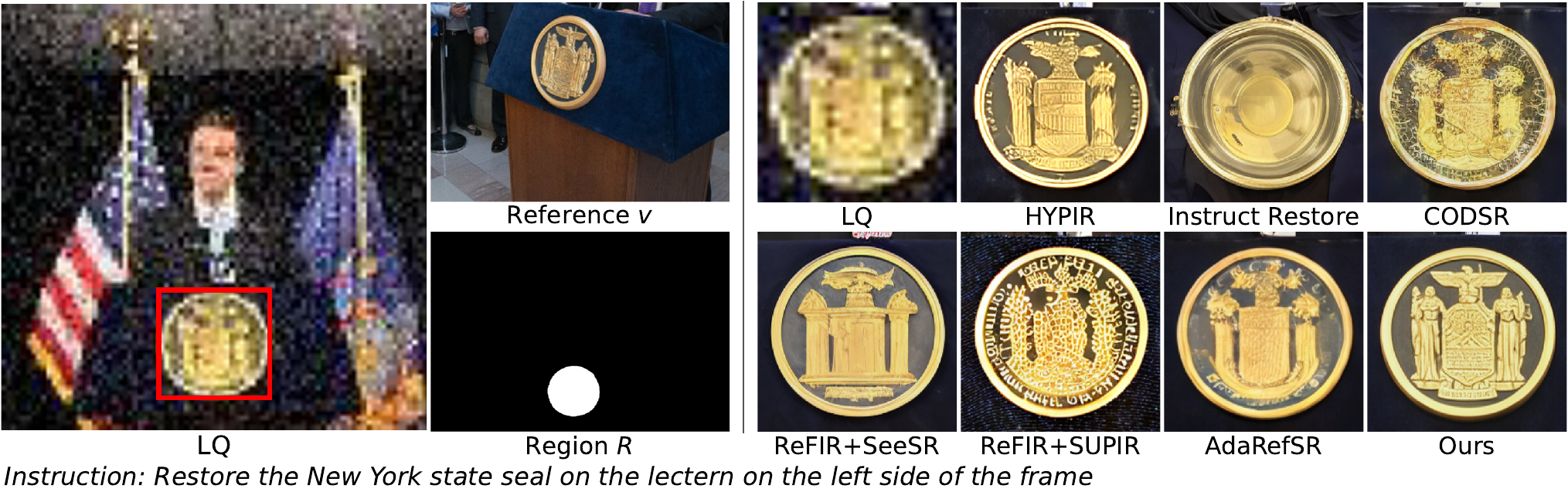}\\
  \includegraphics[width=0.87\linewidth]{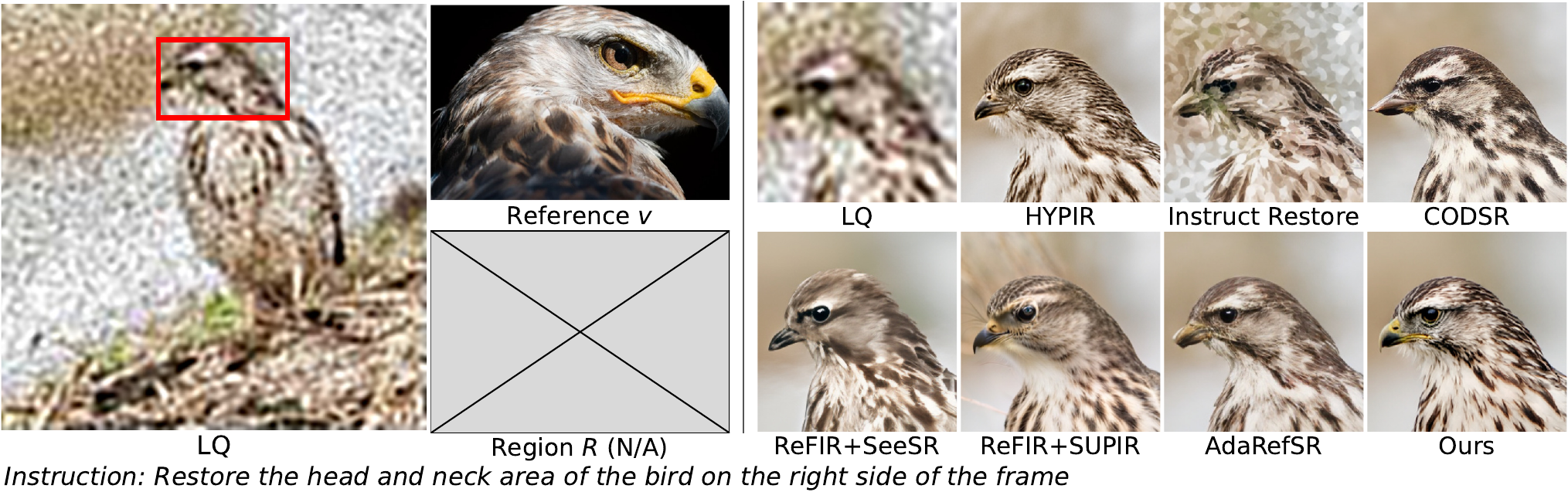}\\
  \caption{\textbf{Qualitative comparison on the proposed test set.} The two condition columns are the visual guidance $v$ and the region control $R$ of Sec.~\ref{sec:method}. FAIR follows the reference inside the target region (red box) while preserving the surroundings.}
  \vspace{-10pt}
  \label{fig:comp_main}
\end{figure*}

\vspace{-4pt}
\subsection{Results}
\label{sec:results}

Among reference-based methods using the same $v$, FAIR reduces FID from
$81.93$ to $63.28$ and LPIPS\textsubscript{in} from $0.0368$ to $0.0319$
(Table~\ref{tab:quant_comparison_synthetic}), indicating better perceptual quality and target-region fidelity. FAIR also achieves the highest DINO-Sim-Ref and CLIP-Sim-Ref scores, showing stronger alignment with $v$ alongside lower GT-based LPIPS\textsubscript{in}. It ranks first on Q-Align and TOPIQ and achieves the lowest LPIPS\textsubscript{in} in every failure tier, with the largest margin on OCR/ID
(Appendix~\ref{app:per_tier}); WR-SR, real-world and runtime results
are in Appendix~\ref{app:experimental_results}.
Fig.~\ref{fig:comp_main} illustrates these differences: competing methods
either leave the region blurred or generate structures inconsistent with the reference, most visibly on the seal, where several baselines invent
an unrelated emblem. A HYPIR-to-FLUX.2-klein-9B cascade that restores first and corrects afterward also underperforms FAIR on every reported metric
(Appendix~\ref{app:cascade}).

FAIR does not lead every metric: AdaRefSR is higher on PSNR and SSIM, and SeeSR-based methods and CODSR on some no-reference scores. PSNR and SSIM reward pixel agreement with a single GT, so a plausible correction of ambiguous detail can lower them~\citep{blau2018perception}, and no-reference scores rate the whole image; regional LPIPS and the blinded study (Sec.~\ref{sec:userstudy}) assess the local repairs directly.

\begin{table}[t]
\centering
\caption{\textbf{Quantitative comparison on the proposed benchmark.} The \textcolor{red}{best} and \textcolor{blue}{second-best} results are highlighted per metric. Reference-similarity metrics (DINO-Sim-Ref and CLIP-Sim-Ref) are reported only for methods that take reference images as input.}
\label{tab:quant_comparison_synthetic}
\renewcommand{\arraystretch}{0.9}
\scriptsize
\vspace{-2pt}
\begin{tabular}{c|ccccc|ccc|c}
\toprule
& \multicolumn{5}{c|}{\textit{Non-reference}} 
& \multicolumn{3}{c|}{\textit{Reference-based}} 
& \textit{Ours} \\
\cmidrule(lr){2-6} \cmidrule(lr){7-9} \cmidrule(lr){10-10}
Metric 
& SeeSR & SUPIR & HYPIR 
& \makecell{Instruct\\Restore} 
& CODSR 
& \makecell{ReFIR\\+ SeeSR} 
& \makecell{ReFIR\\+ SUPIR} 
& AdaRefSR 
& FAIR \\
\midrule
PSNR$\uparrow$       & 20.87 & 20.10 & 20.62 & \textcolor{blue}{20.93} & 20.85 
                     & 20.89 & 19.97 & \textcolor{red}{21.23} & 20.91 \\
SSIM$\uparrow$       & 0.548 & 0.513 & 0.556 & 0.531 & 0.540 
                     & 0.545 & 0.505 & \textcolor{red}{0.561} & \textcolor{blue}{0.558} \\
LPIPS\textsubscript{out}$\downarrow$ & 0.300 & 0.340 & \textcolor{blue}{0.272} & 0.356 & 0.278 
                     & 0.301 & 0.389 & 0.277 & \textcolor{red}{0.243} \\
FID$\downarrow$      & 85.48 & 106.15 & 84.12 & 106.42 & 84.45 
                     & 85.99 & 113.87 & \textcolor{blue}{81.93} & \textcolor{red}{63.28} \\
DISTS$\downarrow$    & 0.1676 & 0.1732 & \textcolor{blue}{0.1511} & 0.2070 & 0.1697 
                     & 0.1676 & 0.1979 & 0.1621 & \textcolor{red}{0.1387} \\
LPIPS\textsubscript{in}$\downarrow$
                     & 0.0406 & 0.0451 & \textcolor{blue}{0.0367} & 0.0466 & 0.0397 
                     & 0.0390 & 0.0489 & 0.0368 & \textcolor{red}{0.0319} \\
\midrule
MUSIQ$\uparrow$      & 72.09 & 70.76 & 70.34 & 68.47 & \textcolor{red}{72.72} 
                     & \textcolor{blue}{72.20} & 68.30 & 69.77 & 71.47 \\
CLIPIQA$\uparrow$    & \textcolor{blue}{0.693} & 0.604 & 0.644 & 0.636 & \textcolor{red}{0.694} 
                     & 0.687 & 0.577 & 0.617 & 0.673 \\
Q-Align$\uparrow$    & 4.382 & 4.467 & 4.340 & 3.903 & \textcolor{blue}{4.483} 
                     & 4.405 & 4.168 & 4.274 & \textcolor{red}{4.514} \\
MANIQA$\uparrow$     & \textcolor{blue}{0.5110} & 0.4306 & 0.4412 & 0.4983 & 0.4434 
                     & \textcolor{red}{0.5123} & 0.4211 & 0.4032 & 0.4356 \\
TOPIQ$\uparrow$      & 0.3978 & 0.3682 & \textcolor{blue}{0.4492} & 0.3732 & 0.4177 
                     & 0.3943 & 0.3448 & 0.4219 & \textcolor{red}{0.4655} \\
\midrule
DINO-Sim-Ref$\uparrow$ & -- & -- & -- & -- & -- 
                     & 0.412 & 0.386 & \textcolor{blue}{0.448} & \textcolor{red}{0.468} \\
CLIP-Sim-Ref$\uparrow$ & -- & -- & -- & -- & -- 
                     & 0.577 & 0.563 & \textcolor{blue}{0.589} & \textcolor{red}{0.598} \\
\bottomrule
\end{tabular}
\vspace{-8pt}
\end{table}

\begin{figure*}[!t]
  \centering
  \begin{minipage}{0.46\textwidth}
    \includegraphics[width=\linewidth]{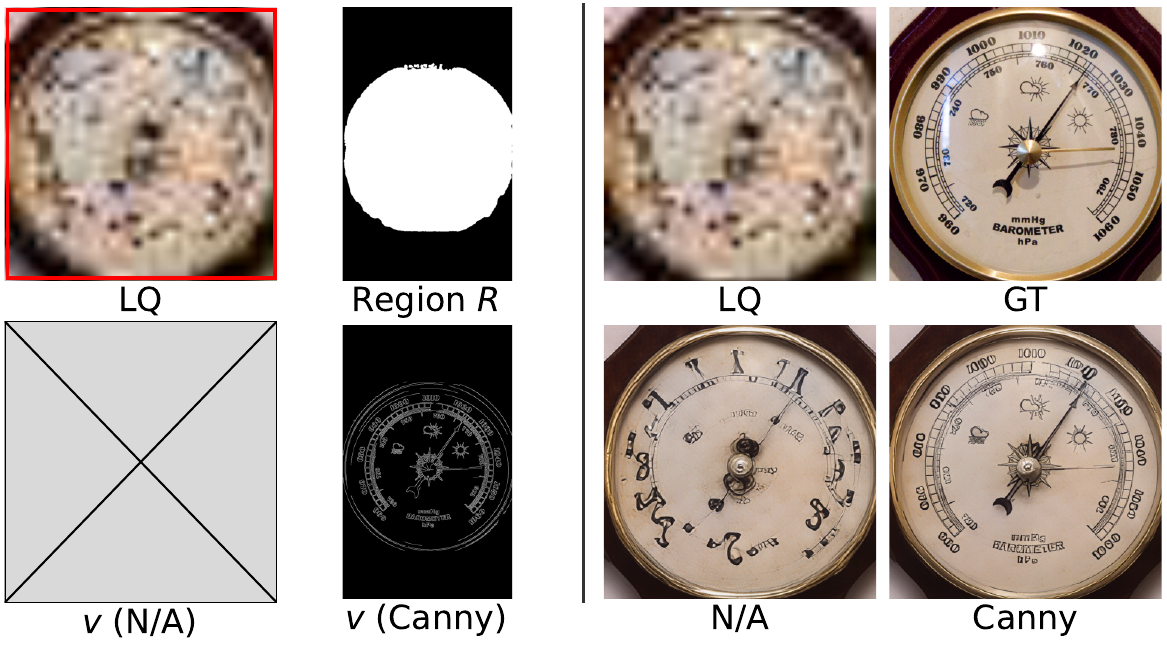}
  \end{minipage}%
  \hspace{0.008\textwidth}%
  \begin{minipage}{0.46\textwidth}
    \includegraphics[width=\linewidth]{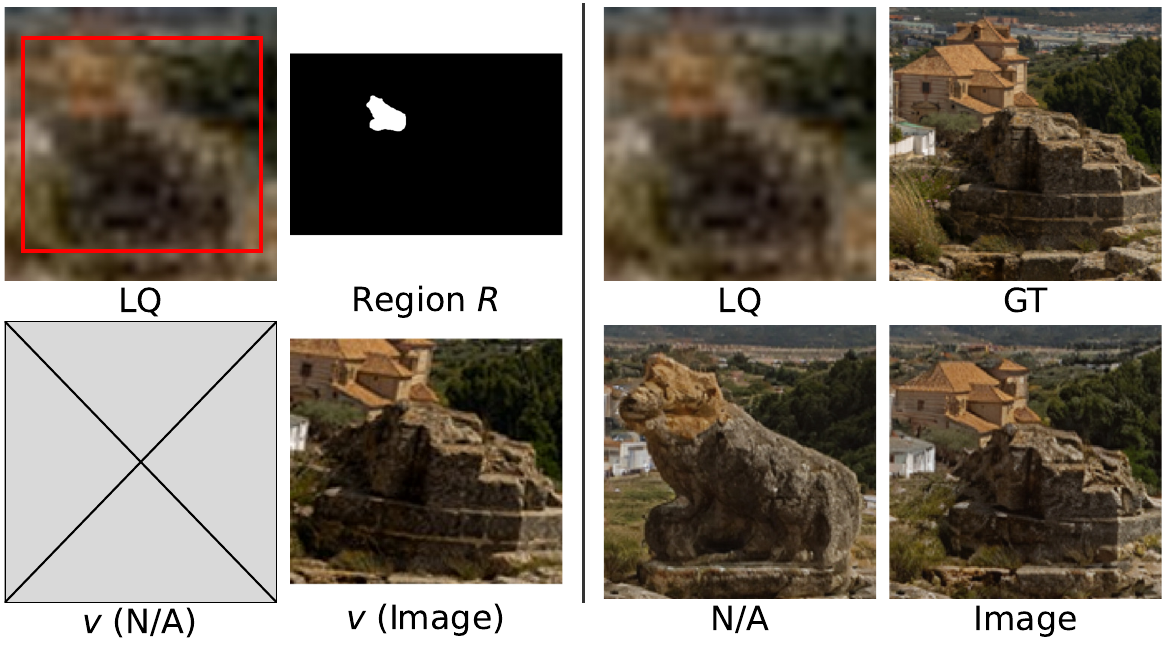}
  \end{minipage}\\[-2pt]
  \begin{minipage}{0.46\textwidth}
    \includegraphics[width=\linewidth]{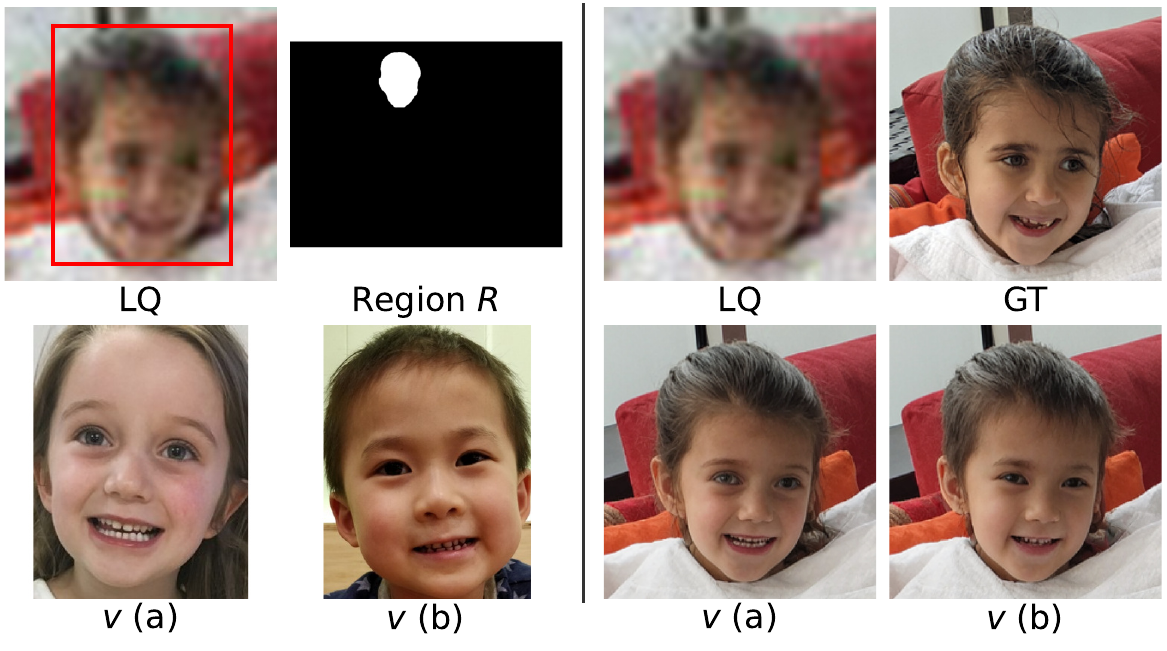}
  \end{minipage}%
  \hspace{0.008\textwidth}%
  \begin{minipage}{0.46\textwidth}
    \includegraphics[width=\linewidth]{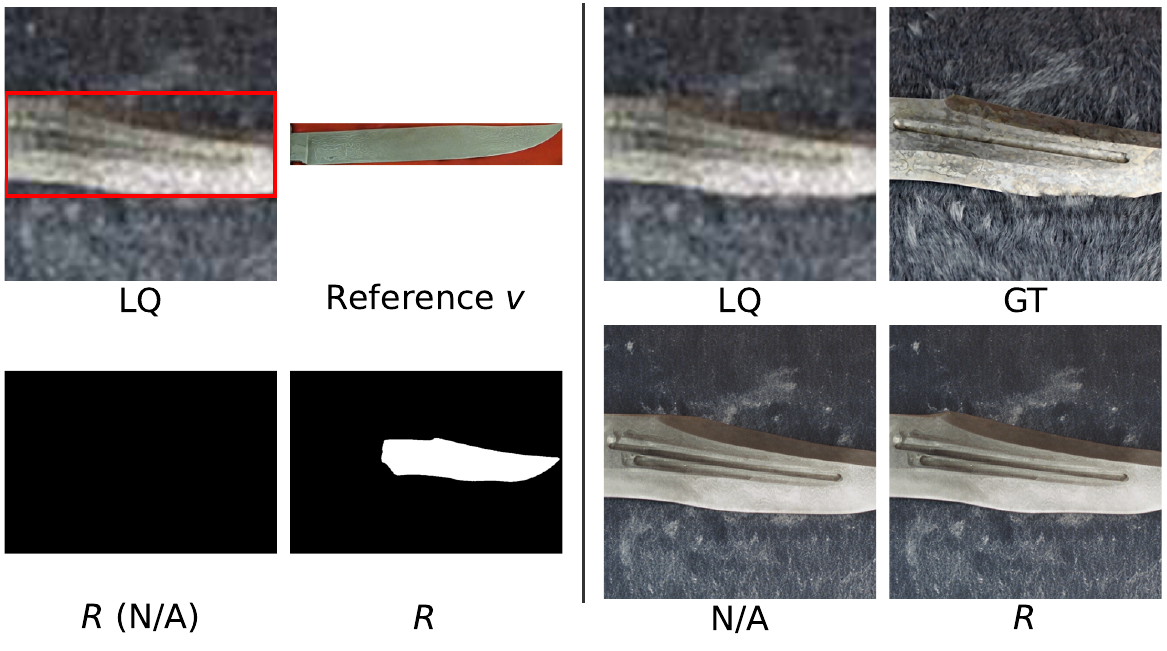}%
  \end{minipage}%
  \vspace{-5pt}
\caption{\textbf{Qualitative ablation of the two conditioning axes.} Restoration guidance $P$ supplies information for recovering target content, while region control $R$ specifies where the guidance applies.}
\vspace{-6pt}
  \label{fig:ablation}
\end{figure*}
\vspace{-4pt}
\subsection{Ablation Study}

\label{sec:ablation}

Table~\ref{tab:ablation} ablates the training design of FAIR: the curriculum, LoRA sharing, joint conditioning, and region-aware reweighting; Fig.~\ref{fig:ablation} and Appendix~\ref{app:condremove} ablate the conditions themselves.
\begin{table}[!t]
\centering
\caption{\textbf{Ablation of the main design choices, using the benchmark's region annotations.}}
\label{tab:ablation}
\renewcommand{\arraystretch}{0.9}
\scriptsize
\vspace{-2pt}
\begin{tabular}{l|cccc}
\toprule
Variant & LPIPS$\downarrow$ & FID$\downarrow$ & MUSIQ$\uparrow$ & CLIPIQA$\uparrow$ \\
\midrule
(a) Stage-1 non-ref restoration only            & 0.326 & 80.38 & 65.02 & 0.578 \\
(b) End-to-end (no curriculum)           & \textcolor{blue}{0.274} & 64.94 & 70.72 & 0.636 \\
(c) Frozen stage-1 LQ LoRA + new Ref LoRA      & 0.285 & 68.57 & 70.45 & 0.626 \\
(d) Joint LQ + ref finetuning Image LoRA (LQ \& Ref share)       & \textcolor{blue}{0.274} & \textcolor{blue}{64.63} & \textcolor{blue}{71.44} & \textcolor{blue}{0.658} \\
\midrule
\textbf{(e) Full \textsc{FAIR}: (d) + reweighting} & \textcolor{red}{0.270} & \textcolor{red}{63.22} & \textcolor{red}{71.50} & \textcolor{red}{0.674} \\
\bottomrule
\end{tabular}
\vspace{2pt}
\end{table}

Variant~(a), stage-1 non-reference restoration, scores worst on all four metrics. End-to-end training~(b) improves performance but trails the two-stage shared-LoRA variants~(d--e) overall. Variant~(c) freezes the stage-1 LQ LoRA and trains a separate reference LoRA, underperforming~(d), which jointly optimizes LQ and reference conditioning in a shared LoRA space. Region-aware reweighting (Eq.~\eqref{eq:rec}) yields the full model~(e), which leads on all four metrics.

Figure~\ref{fig:ablation} shows how $P$ and $R$ affect restoration. A Canny contour
restores the ring of graduations on the barometer dial, and a reference
image restores the masonry of the ruin. Swapping one reference for
another changes the recovered appearance: from the same input, a fair-haired
reference yields lighter hair and grey eyes, a dark-haired one darker
hair and browner eyes. These changes show that $P$ guides the recovered
appearance. Without $R$, the guidance affects content beyond
the target and smooths the blade's lower edge into one curve; with $R$,
the stepped profile survives, showing that the guidance location is
controlled separately.

\vspace{-2pt}
\subsection{Multiple Regions with Per-Region References}

\label{sec:multiregion}

\begin{figure*}[t]
  \centering
  \includegraphics[width=0.95\linewidth]{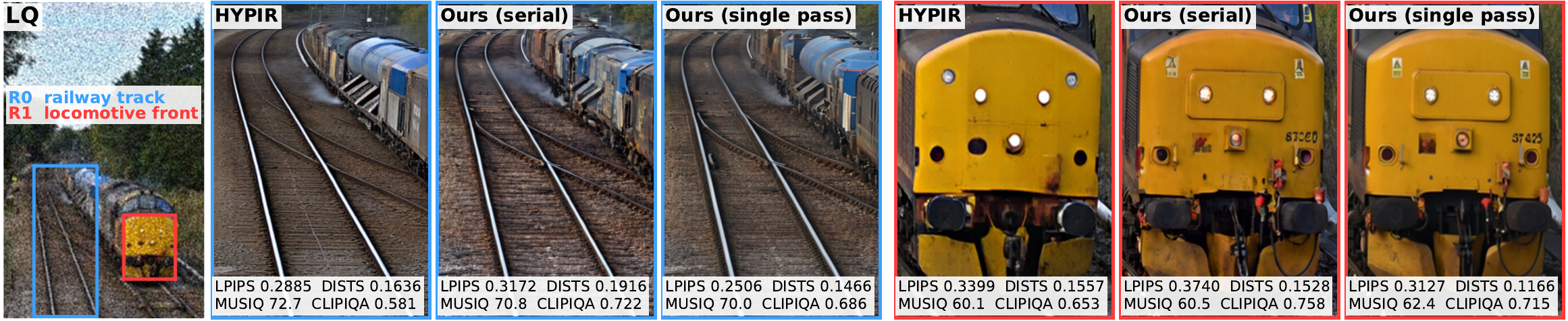}\\[3pt]
  \includegraphics[width=0.95\linewidth]{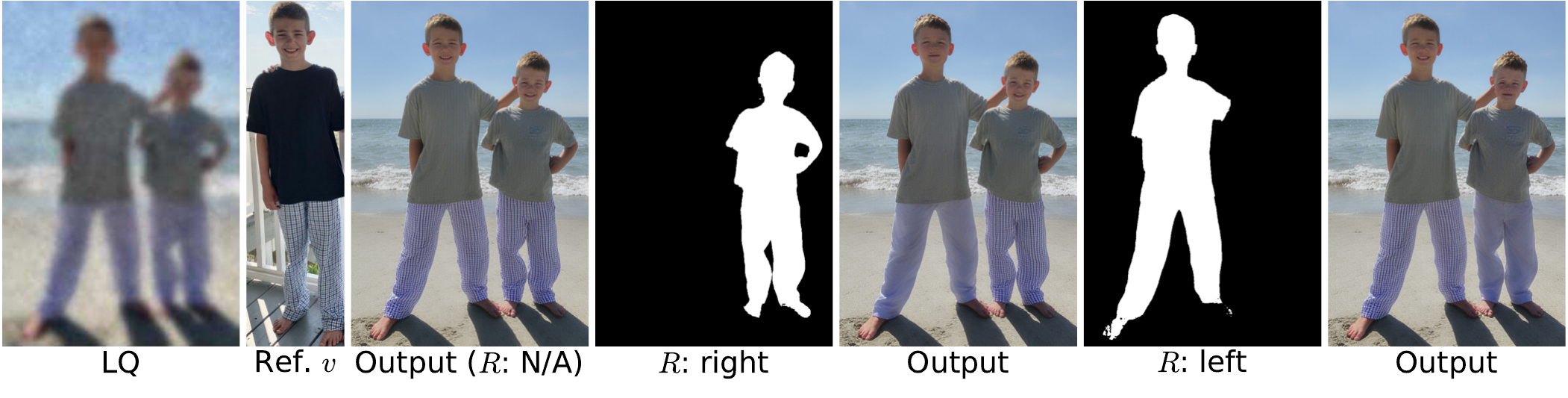}%
  \vspace{-4pt}
  \caption{\textbf{Multi-region and strict region control.} Top: serial and joint restoration of two regions with separate references.
  Bottom: under strict region control, only the boy selected by $R$ follows $v$.}
  \vspace{-4pt}
  \label{fig:multiregion}
\end{figure*}

For an image containing multiple target regions that require different
references, FAIR can restore them jointly. Specifically,
we combine the region masks and tile their corresponding references.
Compared with processing one region per pass, which repeatedly re-encodes the
previous restoration, joint processing avoids the accumulation of artifacts
across passes~\citep{almog2025reedvae}. Figure~\ref{fig:multiregion} compares
the two strategies on target-region crops. Under serial processing, restoring
the second region unintentionally alters the first, whereas joint processing
preserves both. The bottom row shows strict region control: without $R$,
both boys follow the plaid in $v$; with $R$, only the selected boy does.

\vspace{-2pt}
\subsection{Blinded Subjective Study}
\label{sec:userstudy}
\begin{table}[!h]
\centering
\caption{\textbf{User study.} Parentheses give $95\%$ Wilson intervals; HYPIR is second to FAIR.}
\vspace{-4pt}
\label{tab:userstudy}
\footnotesize
\setlength{\tabcolsep}{3pt}
\begin{tabular}{lcc}
\toprule
Dimension & Ours & HYPIR \\
\midrule
Overall quality          & \textbf{79.5\%} \;\scriptsize(73.4--84.5) & 40.0\% \\
Target correction        & \textbf{77.5\%} \;\scriptsize(71.2--82.7) & 44.0\% \\
Non-target preservation    & \textbf{78.5\%} \;\scriptsize(72.3--83.6) & 45.0\% \\
\bottomrule
\end{tabular}
\end{table}

Because IQA metrics may not fully capture human perception, twenty observers
independently selected the top two among nine anonymized outputs for each of
10 cases, evaluating overall quality, target correction, and non-target
preservation separately. FAIR attains $77.5$--$79.5\%$ Top-2 rates compared
with $40.0$--$45.0\%$ for the strongest baseline, HYPIR
(Table~\ref{tab:userstudy}), and ranks first among all methods on each
dimension (Appendix~\ref{app:userstudy}).

\vspace{-2pt}
\subsection{Guidance Settings and Semantic Fidelity}
\label{sec:analysis}

\begin{table}[t]
\centering
\caption{\textbf{Guidance settings at a fixed FLUX.2-klein-9B backbone.} The first four rows vary the guidance source with the final checkpoint fixed (except stage~1); the last three keep $R$ and $t$ of Table~\ref{tab:quant_comparison_synthetic} and change only $v$. DINO similarity is computed between the output's target region and the supplied $v$; content matches use the judge of Appendix~\ref{app:semantic}.}
\label{tab:guidance}
\scriptsize
\renewcommand{\arraystretch}{0.9}
\begin{tabular}{lcccccc}
\toprule
Setting & FID$\downarrow$ & LPIPS\textsubscript{in}$\downarrow$ & LPIPS\textsubscript{out}$\downarrow$ & MUSIQ$\uparrow$ & DINO sim. & Content matches$\uparrow$ \\
\midrule
Stage-1 restoration (Table~\ref{tab:ablation}(a)) & 80.38 & -- & -- & 65.02 & -- & -- \\
Generic ``Restore'' & 71.79 & 0.0353 & 0.2512 & 71.04 & -- & -- \\
Fully LQ-driven guidance & 67.97 & 0.0351 & 0.2486 & 71.21 & -- & -- \\
Annotation-assisted guidance (Table~\ref{tab:quant_comparison_synthetic}) & \textbf{63.28} & \textbf{0.0319} & \textbf{0.2429} & \textbf{71.47} & 0.650 & \textbf{107/120} \\
\midrule
$v$ from another failure category & 68.06 & 0.0349 & 0.2488 & 71.41 & 0.020 & 86/120 \\
$v$ replaced by an unrelated crop & 68.37 & 0.0351 & 0.2490 & 71.45 & 0.030 & 92/120 \\
$v$ omitted & 67.76 & 0.0350 & 0.2484 & 71.16 & -- & 93/120 \\
\bottomrule
\end{tabular}
\end{table}
\textbf{Guidance at a fixed backbone.} In Table~\ref{tab:guidance}, stage-1 restoration on FAIR's backbone reaches FID $80.38$, close to HYPIR ($84.12$) and CODSR ($84.45$). With the final checkpoint fixed, FID falls from $71.79$ with ``Restore'' to $67.97$ with fully LQ-driven guidance and $63.28$ with annotation-assisted guidance; without annotation, FAIR still leads every baseline on FID and LPIPS\textsubscript{in}. HYPIR uses the larger 12B FLUX.1-dev, yet FAIR with ``Restore'' alone has lower FID and LPIPS (Appendix~\ref{app:equalinfo}).

\begin{table}[t]
\centering
\caption{\textbf{Text, identity, and content fidelity on the synthetic benchmark.} CER is averaged over legible text regions, character recall over all text regions, ArcFace similarity over face regions, and content matches over all $120$ regions (evaluated for the three methods shown).}
\label{tab:semantic}
\scriptsize
\renewcommand{\arraystretch}{0.9}
\setlength{\tabcolsep}{3.5pt}
\begin{tabular}{c|ccccc|ccc|c}
\toprule
Metric & SeeSR & SUPIR & HYPIR & \makecell{Instruct\\Restore} & CODSR & \makecell{ReFIR\\+ SeeSR} & \makecell{ReFIR\\+ SUPIR} & AdaRefSR & FAIR \\
\midrule
CER$\downarrow$ & 0.837 & 1.367 & 0.278 & 0.913 & 1.137 & 0.892 & 0.978 & 0.777 & \textbf{0.260} \\
Character recall$\uparrow$ & 30.4\% & 52.4\% & 75.0\% & 42.3\% & 35.5\% & 39.8\% & 18.2\% & 37.2\% & \textbf{81.4\%} \\
ArcFace similarity$\uparrow$ & 0.404 & 0.313 & 0.418 & 0.339 & 0.411 & 0.380 & 0.306 & 0.440 & \textbf{0.605} \\
Content matches$\uparrow$ & -- & -- & 93/120 & -- & -- & -- & -- & 89/120 & \textbf{107/120} \\
\bottomrule
\end{tabular}
\end{table}
\textbf{Wrong or missing references.} Replacing $v$ with a reference from another failure category or an unrelated crop lowers the DINO similarity to it from $0.650$ to $0.02$--$0.03$ and leaves the metrics close to omitting $v$ (Table~\ref{tab:guidance}; full metrics in Table~\ref{tab:wrong_ref}), so FAIR copies little from mismatched references. When retrieval finds no suitable reference, FAIR omits $v$ and still leads all baselines on FID and regional LPIPS (Appendix~\ref{app:robustness}).

\textbf{Text, identity, and content.} Because LPIPS\textsubscript{in} measures perceptual distance, Table~\ref{tab:semantic} checks semantic correctness (Appendix~\ref{app:semantic}). FAIR has the lowest OCR character error rate, the highest character recall and ArcFace identity similarity, and keeps the GT content in $107/120$ regions, against $93$ for HYPIR and $89$ for AdaRefSR.

\vspace{-2pt}
\section{Conclusion}
\label{sec:conclusion}

We address the last-mile problem in generative image restoration, where
localized failures can make an otherwise plausible result unacceptable.
FAIR predicts failure-prone regions from the low-quality input, prepares
restoration guidance $P=(t,v)$ for their needs, and applies it through region
control $R$ during restoration. A 67$k$-region failure-aware dataset,
localized multimodal conditioning, and a two-stage region-aware curriculum
support this approach.
Across synthetic, reference-based, and in-the-wild settings, FAIR improves
perceptual quality, reference alignment, and local controllability. A blinded
study also favors FAIR on overall quality, target correction, and non-target
preservation. These results show the value of identifying where restoration
needs additional evidence and preparing the corresponding guidance in advance.

{
  \bibliographystyle{iclr2027_conference}
  \bibliography{references}
}

\newpage
\appendix

\section*{Appendix Overview}

The appendix is organized as follows. Appendix~\ref{app:method_comparison}
compares FAIR with semantic-guided and reference-based restoration methods.
Appendix~\ref{app:dataset_details} details the three-stage failure-aware data
pipeline and provides additional dataset examples and statistics.
Appendix~\ref{app:training_details} reports the architecture, optimization,
and training configuration. Appendix~\ref{app:experimental_results} presents
additional quantitative results, including a comparison
with a cascade of off-the-shelf models, protocols for text, identity, and content fidelity, inference resolution, lighter degradations, and the cost of each pipeline stage. Appendix~\ref{app:userstudy}
describes the blinded subjective study, and Appendix~\ref{app:robustness}
evaluates automatic condition acquisition, component scale, reference
mismatch, visual-guidance abstention, region control, and run-to-run variation.
Appendix~\ref{app:qualitative_results} collects the additional qualitative comparisons.

\section{Method Comparison}
\label{app:method_comparison}

We compare FAIR with representative diffusion-based restoration
methods along three axes: \emph{auxiliary input} (what external
information is supplied), \emph{region control} (whether and how
that information is spatially localized), and \emph{inference
efficiency} (steps per image). Table~\ref{tab:semantic_methods}
summarizes the comparison.

Semantic-guided
methods~\citep{yang2024pasd, wu2024seesr, yu2024supir, sun2024coser,
liu2025instructrestore, lin2025hypir, chen2025codsr} supply
auxiliary inputs ranging from text prompts and captions to
segmentation masks. Some provide spatial control, but they do not
select restoration guidance by predicting a local failure type.
Reference-based methods~\citep{guo2024refir, yan2025rasr, wang2026adarefsr}
provide external visual evidence, typically without binding it to a
predicted failure region.

FAIR instead predicts failure-prone regions and their types from the
low-quality input, prepares the corresponding restoration guidance,
and applies it through explicit region control.

\begin{table*}[h]
\centering
\caption{\textbf{Comparison of diffusion-based restoration methods.}
Auxiliary input, region control, and inference steps per image.}
\label{tab:semantic_methods}
\scriptsize
\renewcommand{\arraystretch}{0.9}
\begin{tabular}{l|l|l|c}
\toprule
\textbf{Method} & \textbf{Auxiliary Input} & \textbf{Region Control \& Alignment} & \textbf{Steps} \\
\midrule
\multicolumn{4}{l}{\emph{Semantic-guided diffusion restoration}} \\
\midrule
PASD~\citep{yang2024pasd}
  & Detector + caption
  & Implicit (cross-attn)              & 20 \\
SeeSR~\citep{wu2024seesr}
  & Text prompt
  & Implicit (cross-attn)              & 50 \\
SUPIR~\citep{yu2024supir}
  & Text prompt
  & Implicit (cross-attn)              & 50 \\
CoSeR~\citep{sun2024coser}
  & Cognitive embedding
  & Implicit (cross-attn)              & 50 \\
InstructRestore~\citep{liu2025instructrestore}
  & Instruction
  & Instruction-level (implicit)       & 20 \\
HYPIR~\citep{lin2025hypir}
  & Text prompt
  & Implicit (cross-attn)              & 1 \\
CODSR~\citep{chen2025codsr}
  & Tag + segmentation mask
  & Region-adaptive prior + mask       & 1 \\
\midrule
\multicolumn{4}{l}{\emph{Reference-based diffusion restoration}} \\
\midrule
ReFIR~\citep{guo2024refir}
  & Retrieved reference
  & Global (full-image cross-attn)     & 50 \\
RASR~\citep{yan2025rasr}
  & Category-retrieved reference
  & Global (full-image attn)           & 1 \\
AdaRefSR~\citep{wang2026adarefsr}
  & Retrieved reference
  & Adaptive cond.\ (implicit)         & 1 \\
\midrule
\textbf{FAIR}
  & \makecell[l]{\textbf{Multi-modal}\\
      \textbf{(text / reference / contour)}}
  & \makecell[l]{\textbf{Explicit region control (self-attn)}\\
      \textbf{(phrase / bbox / outline / mask)}}
  & \textbf{1} \\
\bottomrule
\end{tabular}
\end{table*}

\section{Data Construction Details}
\label{app:dataset_details}

\begin{table}[h]
\centering
\caption{\textbf{Summary of conditions and notation.} At the task
level, $R$ specifies where additional guidance is applied and $P$ specifies
what guidance is used. At the model input, $P$ is represented by text
tokens $t$ or visual tokens $v$.}
\label{tab:notation}
\scriptsize
\renewcommand{\arraystretch}{0.9}
\setlength{\tabcolsep}{3pt}
\begin{tabular}{@{}llp{0.235\linewidth}p{0.215\linewidth}p{0.155\linewidth}@{}}
\toprule
Symbol & Name & Source & Representation at input & Primarily serves \\
\midrule
$\mathbf{x}_{\mathrm{LQ}}$ & degraded input & degradation of $\mathbf{x}_{\mathrm{GT}}$ & VAE latent, image branch & all tiers \\
$\mathbf{x}_{\mathrm{GT}}$ & clean target & LSDIR, Unsplash, or FFHQ image & supervision only & all tiers \\
$R$ & region control & localization phrase, box, outline, or mask; $\varnothing$ for prompt-only dropout & spatial condition & all tiers \\
$P$ & restoration guidance & text instruction, retrieved reference, or contour & $t$ or $v$ & selected by $F$ \\
$t$ & textual instruction & text instruction $P$ & text-stream tokens & text guidance \\
$v$ & visual guidance & reference or contour $P$ & VAE tokens, image branch & visual guidance \\
$F$ & failure label & fine-tuned VLM & indicates the form of $P$ & --- \\
$\tilde{\mathbf{x}}$ & coarse restoration & existing restoration model & \textbf{offline only} & --- \\
$\hat{\mathbf{x}}$ & restored output & $f_\theta(\mathbf{x}_{\mathrm{LQ}},R,P)$ & model output & all tiers \\
\bottomrule
\end{tabular}
\end{table}

We provide additional details for each stage of the failure-aware 
data construction pipeline described in Sec.~\ref{sec:data}.
The guidance types of Fig.~\ref{fig:valuable_approach} map onto $P=(t,v)$ as
in Table~\ref{tab:notation}: semantic instructions form $t$; reference images,
geometry prompts (Canny contours) and identity correspondences each enter as
$v$; parametric controls, such as the noise and sharpness levels, are
inference settings rather than part of $P$.

\subsection{Stage I: Data Generation and Seed Annotation}

\paragraph{Source images.}
We collect ground-truth images from two sources:
LSDIR~\citep{li2023lsdir}, providing 84,991 high-resolution
natural images curated from Flickr (we use the official training
split), and the Unsplash Lite dataset~\citep{unsplash}, a
curated collection of $\sim$25k photographer-uploaded high-quality
images. All images are resized so that the shorter edge has $1024$
pixels, then center-cropped to $1024 \times 1024$ patches.

\paragraph{Degradation pipeline.}
LQ inputs are synthesized using the second-order degradation pipeline 
of Real-ESRGAN~\citep{wang2021realesrgan}, which sequentially applies 
two rounds of (blur, downsampling, noise, JPEG compression) with 
parameters sampled from the official Real-ESRGAN configuration. 
This pipeline is widely adopted by recent restoration 
methods~\citep{wu2024seesr, wu2024osediff, lin2025hypir} and produces 
LQ images that approximate real-world degradations.

\paragraph{Coarse restorations.}
Each $\mathbf{x}_{\mathrm{LQ}}$ is processed by multiple state-of-the-art
restoration models~\citep{lin2024diffbir,yu2024supir,wu2024seesr,
lin2025hypir}. Their outputs $\tilde{\mathbf{x}}$ expose realistic local
failure modes rather than synthetic perturbations.

\paragraph{Seed annotation prompts.}

We use Qwen3-VL-8B~\citep{bai2025qwen3} with the following prompt structure 
to identify failure regions from
$(\mathbf{x}_{\mathrm{LQ}},\tilde{\mathbf{x}},\mathbf{x}_{\mathrm{GT}})$ triplets.

\paragraph{First-pass prompt (failure localization and typing).}

\begin{quote}\small\ttfamily
You are given three images: a low-quality input (LQ), its restored
result (Restoration), and the corresponding ground-truth image (GT).
Compare the restored result with both the LQ input and GT, and identify
semantically meaningful regions where restoration fails.

For each detected failure, return a JSON object with the following 
fields:
\begin{itemize}\itemsep0pt
  \item \texttt{region\_description}: a brief natural-language 
        description of the region (e.g., "the woman's face on 
        the left", "the text on the street sign").
  \item \texttt{failure\_type}: one of 
        \{OCR/ID, Object Structure, Pattern/Texture, 
        Subjective Feeling\}.
  \item \texttt{severity}: one of \{minor, moderate, severe\}.
  \item \texttt{global\_label}: \texttt{good} if the restored result
        contains no notable local failures relative to the GT;
        \texttt{gt\_skip} if the GT itself is poor quality and
        cannot be used as reference; otherwise \texttt{ok}.
\end{itemize}

If no failures are detected, return an empty list.
\end{quote}

\paragraph{Second-pass prompt (textual re-classification).}

The second pass processes each detected failure independently, 
using only its textual description (without re-accessing the 
images) to re-classify the failure type. This pass corrects 
inconsistencies in the initial typing, particularly for 
borderline cases where Qwen3-VL produced overly broad or 
mis-categorized labels in the first pass.

\begin{quote}\small\ttfamily
Given the following failure region description, re-classify it 
into one of the following failure types: 
\{OCR/ID, Object Structure, Pattern/Texture, Subjective Feeling\}.

Description: \{region\_description\}

Provide your reasoning step by step before giving the final 
classification. Return JSON: 
\{"reasoning": "...", "failure\_type": "..."\}
\end{quote}

\paragraph{Filtering.}
We keep only samples with \texttt{global\_label = ok} and at least one
detected failure.

\paragraph{Manual screening and fine-tuning.}
We manually accept or reject the seed predictions and retain 1K correct
region--type annotations. The verified
$(\mathbf{x}_{\mathrm{LQ}},\tilde{\mathbf{x}},\mathbf{x}_{\mathrm{GT}})$
triplets are then used to fine-tune Qwen3-VL-8B with rank-$64$ LoRA.

\subsection{Stage II: Failure Annotation and Region Construction}

\paragraph{Large-scale offline annotation.}
The fine-tuned annotator processes the full pool using the same input triplet
and predicts a localization phrase, failure label, and validity flag for each
failure. Ground truth and coarse restorations are used only during offline data
construction; the automatic inference mode receives
$\mathbf{x}_{\mathrm{LQ}}$ alone.

\paragraph{Description-to-phrase translation.}

Region descriptions from Stage I are often long, contain Chinese 
fragments, or include qualifiers (e.g., ``the slightly blurred
face of the woman on the right''). For SAM3 to produce a clean 
mask, we first translate each description into a concise English 
noun phrase using Qwen3-VL:

\begin{quote}\small\ttfamily
Convert the following region description into a concise English 
noun phrase suitable for open-vocabulary segmentation. 
The phrase should be 1--6 words and refer unambiguously to 
the object or area.

Description: \{region\_description\}

Output only the noun phrase, with no additional text.
\end{quote}

\paragraph{SAM3 segmentation.}

The translated noun phrase, together with $\mathbf{x}_{\mathrm{GT}}$, is fed into
SAM3~\citep{carion2025sam3} (text-prompted mode) to obtain a binary
mask. We use SAM3's default confidence threshold and retain the mask
with the highest confidence score for each region.

\paragraph{Region filtering and control derivation.}
We filter out regions whose SAM3 mask area falls outside
$[\rho_\text{min},\rho_\text{max}]$ relative to the full image area, using
$\rho_\text{min}=0.5\%$ to remove likely segmentation noise and
$\rho_\text{max}=50\%$ to exclude non-local regions. A bounding box and closed
outline are derived from each retained mask. With the localization phrase,
these instantiate the four region-control formats $R$.

\subsection{Stage III: Reference Retrieval and Guidance Construction}
\label{app:reference_retrieval}

\paragraph{Global reference gallery.}
We partition the LSDIR and Unsplash training images into overlapping
multi-scale tiles at relative scales $\{0.33,0.50,0.67\}$ with stride
ratio $0.5$. DINOv3~\citep{simeoni2025dinov3} features extracted from
these tiles form a global FAISS index containing 4.13M vectors merged
from 18 shards.

\paragraph{Region-aware retrieval and ranking.}
For each annotated region, its local GT crop queries the global index.
We retrieve 1,024 DINO nearest neighbors, exclude candidates from the
same source image, and rerank the top 120 using
\begin{equation}
s = 0.45s_{\mathrm{sem}} + 0.55s_{\mathrm{vis}},
\end{equation}
where $s_{\mathrm{vis}}$ is the DINO cosine similarity to the annotated
local region and $s_{\mathrm{sem}}$ is the CLIP~\citep{radford2021clip}
image--text similarity between the candidate and region description.
We retain candidates with $s\geq0.70$ and
$s_{\mathrm{vis}}\in[0.50,0.98)$; the upper bound removes near-duplicate
photographs.

\paragraph{Semantic recropping and filtering.}
Because grid tiles may truncate the target or contain mostly background,
we prompt SAM3~\citep{carion2025sam3} with the region description on the
candidate source image and recrop the predicted instance box, subject to
a minimum area ratio of $1\%$. SAM3 serves only as a recropper: if no
instance is found, we retain the original tile, since its no-instance
decision is less reliable ($68.4\%$ accuracy) than unconditional
retention ($77.9\%$).
Qwen3-VL-8B then filters candidates for semantic agreement with the
region description, achieving $86.7\%$ validation accuracy. At most one
reference is retained from each source image, yielding three references
from distinct images whenever possible. At inference, if no candidate passes these checks, $v$ is left empty, which is the black-image condition of the reference-slot routing below, and FAIR restores the region from $R$ and $t$ alone (Appendix~\ref{app:robustness}).

\paragraph{Reference-slot routing.}
Across failure tiers, the visual slot of $P$ is routed among a reference
image, a Canny edge map, and black-image dropout with probabilities
$0.760$, $0.068$, and $0.172$, respectively.

\subsection{Dataset Statistics}
\label{app:stats}

Table~\ref{tab:dataset_stats} summarizes the final dataset
composition. The failure-type proportions follow how often each type occurs in the coarse restorations. We keep them so that training matches the failures encountered in practice, and balance the test set across tiers instead (Appendix~\ref{app:per_tier}).

\begin{table}[h]
\centering
\caption{\textbf{Final failure-aware training dataset statistics.} Region
counts are reported after Stage III filtering.}
\label{tab:dataset_stats}
\scriptsize
\renewcommand{\arraystretch}{0.9}
\begin{tabular}{l|r}
\toprule
\textbf{Statistic} & \textbf{Count} \\
\midrule
Stage~I candidate regions       & 169{,}257 \\
Final training regions (post-filter) & 66{,}989 \\
\midrule
\multicolumn{2}{l}{\emph{Failure type distribution}} \\
\midrule
Pattern / Texture & 63.9\% \\
Object Structure   & 21.0\% \\
OCR / ID           & 10.1\% \\
Subjective Feeling & 5.0\% \\
\bottomrule
\end{tabular}
\end{table}

\begin{figure}[t]
    \centering
    \includegraphics[width=\linewidth]{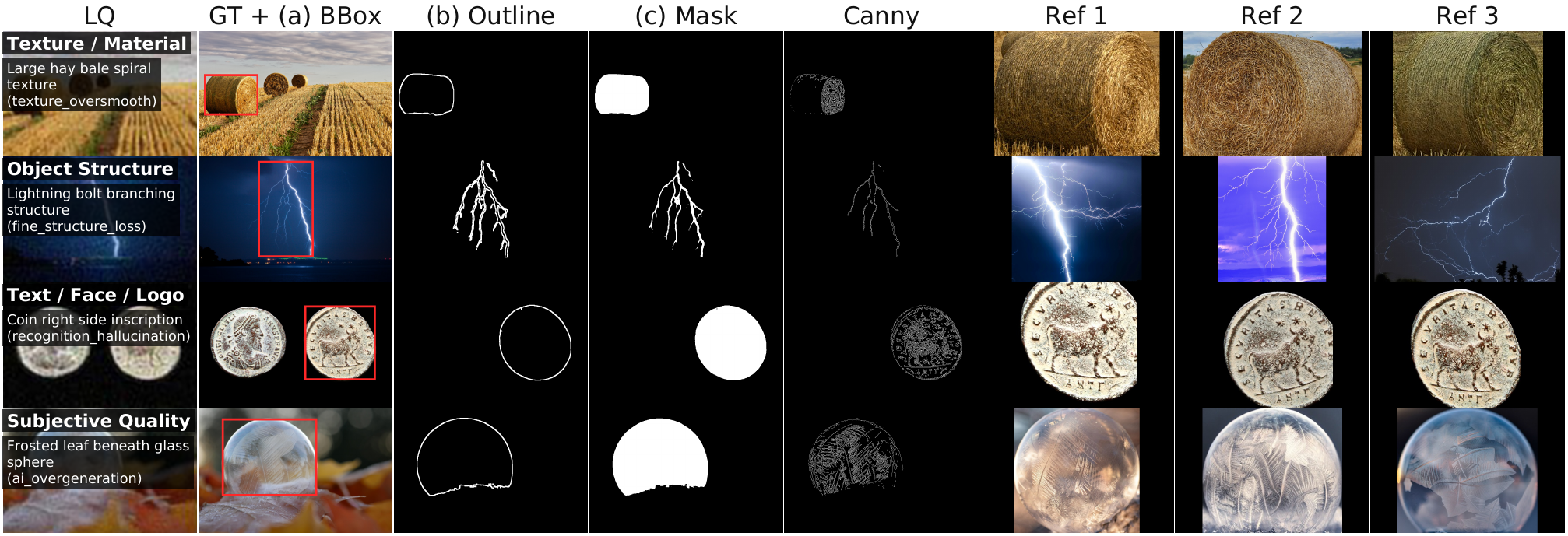}
    \vspace{-8pt}
    \caption{\textbf{Additional samples from our failure-aware dataset, one per tier.} Each row shows the LQ--GT pair, the four region-control formats, the matched $P=(t,v)$, and the failure label $F$.}
    \label{fig:dataset_supp}
    \vspace{-8pt}
\end{figure}

\section{Training and Implementation Details}
\label{app:training_details}

\subsection{Stage 1: Restoration Pre-training}
\label{app:stage1}

Stage~1 establishes a general LQ-to-HQ prior without reference
conditioning. We adapt FLUX.2-klein-9B with LoRA (rank $64$) on
all linear projections per transformer block (text stream frozen),
and train at $1024 \times 1024$ on a three-source mix of Unsplash
regular ($66\%$), Unsplash sharp ($30\%$), and FFHQ ($4\%$),
degraded with the second-order Real-ESRGAN
pipeline~\citep{wang2021realesrgan}. Captions are split
$0.46/0.54$ between the null prompt \texttt{"Restore"} and per-image
LLaVA-style dense captions. Optimization uses RMSprop
($\eta_G = \eta_D = 10^{-6}$) with batch size $64$ across $32$ H200
GPUs (per-GPU batch $2$) for $16{,}000$ iterations. Training uses
fixed-timestep single-step reconstruction at $t = 500$ with noise
augmentation $\sigma_{\text{aug}} \sim \mathcal{U}[200, 800]$: the
predicted velocity is used only to form the one-step clean latent
$z_0 = z_t - \sigma_t v_\theta$, which is decoded to
$\hat{\mathbf{x}}$. No
target velocity is constructed and no separate velocity-supervision
or diffusion loss is applied. The generator loss of
Eq.~\eqref{eq:stage1} therefore acts entirely on the decoded image and
combines image-space $L_2$, LPIPS~\citep{zhang2018lpips}, and a GAN loss
using a ConvNeXt discriminator. We set
$\lambda_{L_2} = \lambda_{\mathrm{LPIPS}} = 10$,
$\lambda_{\mathrm{GAN}} = 0.25$.
The discriminator uses $\mathrm{R1}$ regularization ($\gamma = 100$) enabled
after $1{,}000$ iterations.

\subsection{Stage 2: Region-Guided Fine-Tuning}
\label{app:stage2}

We fine-tune for $30{,}000$ iterations at $512 \times 512$ in
\texttt{bf16} with the same LoRA target set. Each batch mixes
failure-aware data (our $66{,}989$-region dataset with full
reference conditioning) and Real-ESRGAN-degraded restoration data
(Unsplash and FFHQ) in the $4{:}1$ ratio of Sec.~\ref{sec:method}. The conditioning format is sampled as
$\{\texttt{prompt\_only},\texttt{prompt\_loc},\texttt{bbox},
\texttt{outline},\texttt{mask}\}$ with probabilities $\{0.10,0.15,0.25,0.25,0.25\}$. The \texttt{prompt\_only} mode sets $R=\varnothing$, while the localization phrase in \texttt{prompt\_loc} enters $t$ and leaves the binary region map blank. Bounding boxes and outlines are filled to form binary region maps; masks are used directly. The map and $v$ are VAE-encoded and appended to the noisy-target and $\mathbf{x}_{\mathrm{LQ}}$ tokens as separate reference images, each with its own positional offset, so that every condition reaches the target through the backbone's joint self-attention over a single token stream. The visual guidance $v$
follows the reference-slot routing in
Appendix~\ref{app:reference_retrieval}. The training loss
follows Stage~1 with $\lambda_{L_2} = \lambda_{\mathrm{LPIPS}} = 10$
and $\lambda_{\mathrm{GAN}} = 0.25$, plus the region-aware reweighting
$w = 1 + 4(1 - m)$ of Eq.~\eqref{eq:rec}. The reweighting is applied
to the image-space $L_2$ term \emph{only}; LPIPS and the adversarial
loss remain image-wide, since spatially masking them would weaken the
perceptual and realism constraints in the target region. As in stage~1,
all losses act on the decoded
image. Optimization keeps RMSprop ($\eta_G = \eta_D = 10^{-6}$) as in stage~1.

\section{Additional Experimental Results}
\label{app:experimental_results}

\subsection{Real-World and WR-SR Benchmarks}

The synthetic-set findings of Sec.~\ref{sec:results} carry over to
both real-world settings. In Tables~\ref{tab:quant_comparison_real}--\ref{tab:equalinfo} and~\ref{tab:cascade}, the best and
second-best results are shown in \textcolor{red}{red} and \textcolor{blue}{blue}.
On the in-the-wild real-world test set of $207$ images~\citep{yin2026far},
shown in Table~\ref{tab:quant_comparison_real}, FAIR is best on both
reference-similarity metrics and second on MUSIQ, CLIPIQA, and Q-Align.
CODSR and InstructRestore score higher on some no-reference metrics.
Nano~Banana, a general-purpose commercial image model, scores lower
on the reported metrics.
On the $80$ WR-SR test images~\citep{jiang2021wrsr}, shown in
Table~\ref{tab:quant_comparison}, FAIR is best on five of
ten metrics, SSIM, LPIPS, FID and both reference-side similarities,
the latter by wide margins ($+0.056$ DINO, $+0.030$ CLIP). PSNR is
within $0.24$~dB of AdaRefSR, while CODSR and the SeeSR-based methods
lead the no-reference scores. Runtime, reported in the last row of
Table~\ref{tab:quant_comparison}, is $3.16$~s, competitive with AdaRefSR ($2.30$~s). Figure~\ref{fig:comp_appx_ext} shows qualitative results on both sets.

\textbf{Cost of the full pipeline.} FAIR restores in one pass. Users can supply $R$, $t$, and $v$ directly and run only the restorer; otherwise the automatic stages of Table~\ref{tab:stage_cost} supply them. Detection uses merged LoRA weights and a static KV cache. Retrieval keeps the index of $4.1$M DINOv3-L/16 tile features on the GPU (\texttt{search\_k}$\,{=}\,64$), reranks with CLIP-B/32 (top eight), and recrops and verifies the selected candidate with SAM3 and Qwen3-VL-4B. Per query, the search takes $3$~ms, candidate loading $0.11$~s, recropping and verification $0.06$~s each, and embedding and reranking $0.02$~s. Each stage is timed separately. At one query per image, the summed stage times add $1.08$~s to the $3.16$~s restoration measured on WR-SR, for an estimated $4.24$~s per image, about $26\%$ of it for guidance.

\begin{table}[!ht]
\centering
\caption{\textbf{Per-stage cost of the automatic pipeline.} Restoration time is measured on WR-SR; the condition stages are optional when users supply $R$ and $P$.}
\label{tab:stage_cost}
\scriptsize
\renewcommand{\arraystretch}{0.9}
\begin{tabular}{llcc}
\toprule
Stage & Model parameters & Recorded peak GPU memory & Time per image \\
\midrule
Failure detection & 8.77B (Qwen3-VL-8B) & 20.5 GB & 0.76 s \\
Region segmentation & 0.86B (SAM3) & -- & 0.07 s \\
Retrieval and verification & 0.45B (DINOv3 and CLIP) & 21.5 GB & $\approx$0.25 s \\
FAIR restoration & $\approx$17B (DiT and text encoder) & 41 GB & 3.16 s \\
\bottomrule
\end{tabular}
\end{table}

\begin{table}[!t]
\centering
\caption{\textbf{Quantitative comparison on the real-world test set.}}
\label{tab:quant_comparison_real}
\setlength{\tabcolsep}{3pt}
\scriptsize
\renewcommand{\arraystretch}{0.9}
\begin{tabular}{c|cccccc|ccc|c}
\toprule
& \multicolumn{6}{c|}{\textit{Non-reference}} 
& \multicolumn{3}{c|}{\textit{Reference-based}} 
& \textit{Ours} \\
\cmidrule(lr){2-7} \cmidrule(lr){8-10} \cmidrule(lr){11-11}
Metric 
& SeeSR & SUPIR & HYPIR 
& \makecell{Instruct\\Restore} 
& CODSR 
& \makecell{Nano\\Banana} 
& \makecell{ReFIR\\+ SeeSR} 
& \makecell{ReFIR\\+ SUPIR} 
& AdaRefSR 
& FAIR \\
\midrule
MUSIQ$\uparrow$       & 66.01 & 56.22 & 64.72 & 65.06 & \textcolor{red}{69.93} & 45.99
                      & 66.07 & 55.70 & 59.65 & \textcolor{blue}{69.62} \\
CLIPIQA$\uparrow$     & 0.633 & 0.433 & 0.601 & \textcolor{red}{0.686} & 0.653 & 0.402
                      & 0.628 & 0.428 & 0.514 & \textcolor{blue}{0.671} \\
Q-Align$\uparrow$     & 3.803 & 3.535 & 3.883 & 3.706 & \textcolor{red}{4.171} & 3.339
                      & 3.815 & 3.445 & 3.727 & \textcolor{blue}{4.068} \\
\midrule
DINO-Sim-Ref$\uparrow$ & -- & -- & -- & -- & -- & --
                       & 0.363 & \textcolor{blue}{0.364} & 0.363 & \textcolor{red}{0.369} \\
CLIP-Sim-Ref$\uparrow$ & -- & -- & -- & -- & -- & --
                       & 0.531 & 0.533 & \textcolor{blue}{0.534} & \textcolor{red}{0.542} \\
\bottomrule
\end{tabular}
\end{table}

\begin{table}[!t]
\centering
\caption{\textbf{Quantitative comparison on the WR-SR test set.}}
\label{tab:quant_comparison}
\scriptsize
\renewcommand{\arraystretch}{0.9}
\begin{tabular}{c|ccccc|ccc|c}
\toprule
& \multicolumn{5}{c|}{\textit{Non-reference}} 
& \multicolumn{3}{c|}{\textit{Reference-based}} 
& \textit{Ours} \\
\cmidrule(lr){2-6} \cmidrule(lr){7-9} \cmidrule(lr){10-10}
Metric 
& SeeSR & SUPIR & HYPIR 
& \makecell{Instruct\\Restore} 
& CODSR 
& \makecell{ReFIR\\+ SeeSR} 
& \makecell{ReFIR\\+ SUPIR} 
& AdaRefSR 
& FAIR \\
\midrule
PSNR$\uparrow$       & 21.72 & 20.77 & 21.10 & 21.24 & 21.42 
                     & \textcolor{blue}{21.85} & 21.04 & \textcolor{red}{22.04} & 21.80 \\
SSIM$\uparrow$       & 0.570 & 0.528 & 0.565 & 0.558 & 0.547 
                     & 0.569 & 0.539 & \textcolor{blue}{0.581} & \textcolor{red}{0.591} \\
LPIPS$\downarrow$    & 0.349 & 0.394 & 0.335 & 0.430 & 0.333 
                     & 0.341 & 0.394 & \textcolor{blue}{0.298} & \textcolor{red}{0.262} \\
FID$\downarrow$      & 64.64 & 85.50 & 76.93 & 121.71 & 62.20 
                     & 61.22 & 77.55 & \textcolor{blue}{48.85} & \textcolor{red}{38.90} \\
\midrule
MUSIQ$\uparrow$      & 72.09 & 69.94 & 69.66 & 69.53 & \textcolor{red}{72.92} 
                     & \textcolor{blue}{72.42} & 70.06 & 70.21 & 71.37 \\
CLIPIQA$\uparrow$    & 0.711 & 0.624 & 0.664 & 0.606 & \textcolor{red}{0.724} 
                     & \textcolor{blue}{0.715} & 0.623 & 0.669 & 0.703 \\
Q-Align$\uparrow$    & 4.40 & 4.30 & 4.19 & 3.79 & \textcolor{red}{4.58} 
                     & 4.44 & 4.30 & 4.36 & \textcolor{blue}{4.48} \\
\midrule
DINO-Sim-Ref$\uparrow$ & -- & -- & -- & -- & -- 
                       & 0.689 & 0.663 & \textcolor{blue}{0.768} & \textcolor{red}{0.824} \\
CLIP-Sim-Ref$\uparrow$ & -- & -- & -- & -- & -- 
                       & 0.750 & 0.727 & \textcolor{blue}{0.782} & \textcolor{red}{0.811} \\
\midrule
Runtime/s$\downarrow$ & 34.91 & 37.61 & 5.76 & 17.58 & \textcolor{red}{1.71} 
                      & 35.06 & 40.68 & \textcolor{blue}{2.30} & 3.16 \\
\bottomrule
\end{tabular}
\end{table}

\subsection{Benchmark Composition}
\label{app:benchmark}

The synthetic benchmark holds $120$ triplets, $30$ per failure tier,
built by the Sec.~\ref{sec:data} pipeline; none of its regions is among
the $66{,}989$ training regions. References are constructed by the same
retrieval procedure as in training
(Appendix~\ref{app:reference_retrieval}). LPIPS\textsubscript{in} and LPIPS\textsubscript{out} are computed on
the full image after replacing the pixels outside or inside the GT failure
mask, respectively, with GT. DINO- and CLIP-Sim-Ref measure how closely the output follows $v$, whereas LPIPS\textsubscript{in} measures perceptual distance to GT inside the region; a well-matched reference can bring the region closer to both, and shuffling references worsens both (Table~\ref{tab:ref_shuffle}). At inference, the long image side is capped at 1024\,px without upscaling on the synthetic and real-world sets, and WR-SR inputs are used at their native 480--1024\,px. Outputs are scored against $\mathbf{x}_{\mathrm{GT}}$ at its native resolution; Appendix~\ref{app:resolution} compares inference at 512 and 1024\,px.

\paragraph{Conditions at test time.}
Guidance $(R,t,v)$ can be user-supplied or prepared by the pipeline. On the synthetic benchmark, $R$ is predicted from $\mathbf{x}_{\mathrm{LQ}}$ (Appendix~\ref{app:robustness}); annotated instructions $t$ (Sec.~\ref{sec:data}) and references $v$ retrieved with clean GT region crops (Appendix~\ref{app:reference_retrieval}) simulate user-provided repair requests and references. FAIR and reference-based baselines receive the same $v$. WR-SR supplies paired references $v$, with whole-image $R$ and a Qwen3-VL caption of $\mathbf{x}_{\mathrm{LQ}}$ as $t$. On the real-world set without GT, $R$, $t$, and the reference-retrieval query crop all come from $\mathbf{x}_{\mathrm{LQ}}$, using the gallery of Appendix~\ref{app:reference_retrieval}. Two annotation-free synthetic settings are also evaluated (Sec.~\ref{sec:analysis}): fully LQ-driven guidance uses the same input sources as the real-world setting, while generic restoration receives only $\mathbf{x}_{\mathrm{LQ}}$ and the instruction ``Restore''.

\subsection{Results by Failure Tier}
\label{app:per_tier}

The test set holds $30$ regions per tier, so the aggregate numbers of
Table~\ref{tab:quant_comparison_synthetic} are not dominated by the
tier that is most frequent in training (Pattern/Texture, $63.9\%$ of
regions). Table~\ref{tab:per_tier} breaks LPIPS\textsubscript{in} down
by tier. FAIR leads every tier, and by the widest margin on OCR/ID, the
identity-critical case that motivates the taxonomy: $0.0103$ against
$0.0172$ for the best baseline, a $40\%$ reduction, against $7\%$ on
Pattern/Texture. Outside the target, FAIR also has the lowest
LPIPS\textsubscript{out} in every tier, so the in-region gains do not come
at the expense of the rest of the image.

\begin{table}[h]
\centering
\caption{\textbf{\boldmath LPIPS\textsubscript{in} and LPIPS\textsubscript{out} by failure tier
($30$ regions each).}}
\label{tab:per_tier}
\scriptsize
\renewcommand{\arraystretch}{0.9}
\setlength{\tabcolsep}{6pt}
\begin{tabular}{lccccc}
\toprule
Tier & ReFIR+SeeSR & HYPIR & CODSR & AdaRefSR & \textsc{FAIR} \\
\midrule
\multicolumn{6}{l}{\emph{LPIPS\textsubscript{in}}$\downarrow$} \\
OCR / ID             & 0.0190 & 0.0188 & 0.0203 & \textcolor{blue}{0.0172} & \textcolor{red}{0.0103} \\
Object Structure     & 0.0331 & 0.0303 & 0.0317 & \textcolor{blue}{0.0302} & \textcolor{red}{0.0264} \\
Pattern / Texture    & 0.0611 & \textcolor{blue}{0.0573} & 0.0624 & 0.0586 & \textcolor{red}{0.0534} \\
Subjective Feeling   & 0.0430 & \textcolor{blue}{0.0403} & 0.0443 & 0.0411 & \textcolor{red}{0.0377} \\
\midrule
\multicolumn{6}{l}{\emph{LPIPS\textsubscript{out}}$\downarrow$} \\
OCR / ID             & 0.306 & \textcolor{blue}{0.279} & 0.283 & 0.280 & \textcolor{red}{0.238} \\
Object Structure     & 0.286 & 0.272 & 0.275 & \textcolor{blue}{0.267} & \textcolor{red}{0.242} \\
Pattern / Texture    & 0.302 & 0.265 & \textcolor{blue}{0.263} & 0.278 & \textcolor{red}{0.247} \\
Subjective Feeling   & 0.309 & \textcolor{blue}{0.270} & 0.293 & 0.284 & \textcolor{red}{0.245} \\
\bottomrule
\end{tabular}
\end{table}

\subsection{Equal-Information Comparison}
\label{app:equalinfo}

To separate the model from the extra conditions, we run FAIR with
$\mathbf{x}_{\mathrm{LQ}}$ alone: $R$ and $v$ are left empty and the
instruction is the generic ``Restore''; checkpoint and inference settings
match Table~\ref{tab:quant_comparison_synthetic}. Against the five baselines
without $v$ (Table~\ref{tab:equalinfo}), FAIR still leads six of eleven
metrics, including LPIPS\textsubscript{out}, LPIPS\textsubscript{in}, FID
and DISTS, so its advantage does not come from the conditions alone. Adding
$R$, $t$ and $v$ then lowers FID from $71.79$ to $63.28$ and
LPIPS\textsubscript{in} from $0.0353$ to $0.0319$.

\begin{table}[h]
\centering
\caption{\textbf{Equal-information comparison on the synthetic benchmark.}
\textsc{FAIR} (LQ only) receives $\mathbf{x}_{\mathrm{LQ}}$ alone;
\textcolor{red}{best} and \textcolor{blue}{second-best} among the methods
without $v$ are highlighted. The last column repeats \textsc{FAIR} with its
conditions (Table~\ref{tab:quant_comparison_synthetic}).}
\label{tab:equalinfo}
\scriptsize
\renewcommand{\arraystretch}{0.9}
\begin{tabular}{c|ccccc|c|c}
\toprule
Metric & SeeSR & SUPIR & HYPIR & \makecell{Instruct\\Restore} & CODSR
& \makecell{\textsc{FAIR}\\(LQ only)} & \textsc{FAIR} \\
\midrule
PSNR$\uparrow$ & \textcolor{blue}{20.87} & 20.10 & 20.62 & \textcolor{red}{20.93} & 20.85 & 20.80 & 20.91 \\
SSIM$\uparrow$ & 0.548 & 0.513 & \textcolor{red}{0.556} & 0.531 & 0.540 & \textcolor{blue}{0.551} & 0.558 \\
LPIPS\textsubscript{out}$\downarrow$ & 0.300 & 0.340 & \textcolor{blue}{0.272} & 0.356 & 0.278 & \textcolor{red}{0.251} & 0.243 \\
FID$\downarrow$ & 85.48 & 106.15 & \textcolor{blue}{84.12} & 106.42 & 84.45 & \textcolor{red}{71.79} & 63.28 \\
DISTS$\downarrow$ & 0.1676 & 0.1732 & \textcolor{blue}{0.1511} & 0.2070 & 0.1697 & \textcolor{red}{0.1495} & 0.1387 \\
LPIPS\textsubscript{in}$\downarrow$ & 0.0406 & 0.0451 & \textcolor{blue}{0.0367} & 0.0466 & 0.0397 & \textcolor{red}{0.0353} & 0.0319 \\
\midrule
MUSIQ$\uparrow$ & \textcolor{blue}{72.09} & 70.76 & 70.34 & 68.47 & \textcolor{red}{72.72} & 71.04 & 71.47 \\
CLIPIQA$\uparrow$ & \textcolor{blue}{0.693} & 0.604 & 0.644 & 0.636 & \textcolor{red}{0.694} & 0.663 & 0.673 \\
Q-Align$\uparrow$ & 4.382 & 4.467 & 4.340 & 3.903 & \textcolor{blue}{4.483} & \textcolor{red}{4.485} & 4.514 \\
MANIQA$\uparrow$ & \textcolor{red}{0.5110} & 0.4306 & 0.4412 & \textcolor{blue}{0.4983} & 0.4434 & 0.4249 & 0.4356 \\
TOPIQ$\uparrow$ & 0.3978 & 0.3682 & \textcolor{blue}{0.4492} & 0.3732 & 0.4177 & \textcolor{red}{0.4557} & 0.4655 \\
\bottomrule
\end{tabular}
\end{table}

\subsection{Removing the Visual Guidance at Test Time}
\label{app:condremove}

Removing $v$ while keeping $R$ and $t$
raises LPIPS\textsubscript{in} by $68\%$ on OCR/ID but only $1.6\%$ on
Subjective Feeling (Table~\ref{tab:vbytier}). The harder the tier, the more
the external evidence matters, as the taxonomy of
Fig.~\ref{fig:valuable_approach} assumes; Subjective Feeling, which it serves
with an instruction alone, barely needs a reference.

\begin{table}[h]
\centering
\caption{\textbf{\boldmath LPIPS\textsubscript{in} with and without $v$ by failure
tier ($30$ regions each).}}
\label{tab:vbytier}
\scriptsize
\renewcommand{\arraystretch}{0.9}
\setlength{\tabcolsep}{2pt}
\begin{tabular}{l|ccc}
\toprule
Tier & with $v$ & without $v$ & $p$ \\
\midrule
OCR / ID           & 0.0103 & 0.0173 \scriptsize($+68\%$) & $4\times10^{-6}$ \\
Object Structure   & 0.0264 & 0.0291 \scriptsize($+10\%$) & $0.022$ \\
Pattern / Texture  & 0.0534 & 0.0553 \scriptsize($+3.6\%$) & $0.047$ \\
Subjective Feeling & 0.0377 & 0.0383 \scriptsize($+1.6\%$) & $0.22$ \\
\bottomrule
\end{tabular}
\end{table}

\subsection{Text, Identity, and Content Fidelity}
\label{app:semantic}

LPIPS\textsubscript{in} measures perceptual distance, so we also check whether the restored regions keep the right text, identity, and content (Table~\ref{tab:semantic}).
\textbf{Text.} Qwen3-VL-4B transcribes each text region, and we compute the character error rate (CER) against the annotated text. CER is averaged over the regions whose clean GT crop the same reader transcribes with CER $\leq 0.35$, so that illegible ground truth does not dominate; character recall, the fraction of GT characters recovered, is averaged over all annotated text regions.
\textbf{Identity.} RetinaFace detects landmarks on each GT face crop, every method's output is aligned with the same transform, and we report the ArcFace cosine similarity to GT.
\textbf{Content.} For each of the $120$ regions, Qwen3-VL-4B first identifies the GT content (object structure, text, identity, logos, or material patterns) and then checks whether the output keeps it; less sharp outputs count as matches when the content is preserved. Responses are limited to $220$ tokens, and a response without a final verdict counts as a non-match. FAIR matches $107/120$ regions, against $93/120$ for HYPIR, $89/120$ for AdaRefSR, and $66/120$ for the LQ input. The gap is largest on OCR/ID ($28$ against $23$ for HYPIR) and Object Structure ($26$ against $21$).
\subsection{Inference Resolution}
\label{app:resolution}

Stage~2 fine-tunes at $512^2$, whereas inference caps the long side at 1024\,px. We run the Table~\ref{tab:quant_comparison_synthetic} setting with the same model and the long side capped at 512 or 1024\,px, and score both outputs after resizing them to a 512\,px long side (Table~\ref{tab:resolution}). In-region LPIPS is nearly identical ($0.0235$ and $0.0237$), so target-region fidelity does not depend on matching the training resolution. The 1024\,px cap scores better on out-of-region LPIPS and PSNR, DINO similarity to $v$, and both no-reference metrics.

\begin{table}[h]
\centering
\caption{\textbf{Inference at a 512 or 1024\,px long-side cap}, both scored at a 512\,px long side. DINO and CLIP similarities are computed between the output's target-region crop and $v$.}
\label{tab:resolution}
\scriptsize
\renewcommand{\arraystretch}{0.9}
\setlength{\tabcolsep}{3.5pt}
\begin{tabular}{lccccccc}
\toprule
Inference cap & LPIPS\textsubscript{in}$\downarrow$ & LPIPS\textsubscript{out}$\downarrow$ & PSNR\textsubscript{out}$\uparrow$ & DINO sim.$\uparrow$ & CLIP sim.$\uparrow$ & MUSIQ$\uparrow$ & CLIP-IQA$\uparrow$ \\
\midrule
512\,px & 0.0235 & 0.1671 & 22.39 & 0.606 & 0.650 & 69.9 & 0.688 \\
1024\,px & 0.0237 & 0.1629 & 22.64 & 0.622 & 0.643 & 71.8 & 0.735 \\
\bottomrule
\end{tabular}
\end{table}

\subsection{Lighter Degradations}
\label{app:light}

We also test inputs that are easier to restore. We degrade the same $120$ images at three milder levels of the benchmark degradation (milder downsampling, blur, compression, and noise), with input PSNR of $31.0$, $27.0$, and $24.2$\,dB, and run the full automatic pipeline, deriving $R$, $t$, and $v$ from $\mathbf{x}_{\mathrm{LQ}}$. HYPIR uses the settings of Table~\ref{tab:quant_comparison_synthetic}, and AdaRefSR receives the references of Table~\ref{tab:quant_comparison_synthetic}. At every level, FAIR has the lowest LPIPS, DISTS, FID, and regional LPIPS (Table~\ref{tab:light}), and it has the highest PSNR at the two lighter levels.

\begin{table}[h]
\centering
\caption{\textbf{Results at three lighter degradation levels} (fully automatic guidance for FAIR).}
\label{tab:light}
\scriptsize
\renewcommand{\arraystretch}{0.9}
\begin{tabular}{llccccc}
\toprule
Input PSNR & Method & PSNR$\uparrow$ & LPIPS$\downarrow$ & DISTS$\downarrow$ & FID$\downarrow$ & LPIPS\textsubscript{in} / LPIPS\textsubscript{out}$\downarrow$ \\
\midrule
\multirow{3}{*}{31.0\,dB} & HYPIR & 23.22 & 0.166 & 0.105 & 44.1 & 0.0201 / 0.151 \\
& AdaRefSR & 24.13 & 0.156 & 0.093 & 27.4 & 0.0180 / 0.140 \\
& FAIR & \textbf{24.85} & \textbf{0.110} & \textbf{0.070} & \textbf{23.6} & \textbf{0.0127} / \textbf{0.100} \\
\midrule
\multirow{3}{*}{27.0\,dB} & HYPIR & 22.99 & 0.170 & 0.104 & 43.7 & 0.0201 / 0.154 \\
& AdaRefSR & 23.75 & 0.183 & 0.103 & 34.4 & 0.0212 / 0.164 \\
& FAIR & \textbf{24.10} & \textbf{0.140} & \textbf{0.083} & \textbf{28.8} & \textbf{0.0163} / \textbf{0.126} \\
\midrule
\multirow{3}{*}{24.2\,dB} & HYPIR & 22.11 & 0.213 & 0.117 & 53.4 & 0.0247 / 0.193 \\
& AdaRefSR & \textbf{23.08} & 0.229 & 0.123 & 44.2 & 0.0263 / 0.206 \\
& FAIR & 22.60 & \textbf{0.196} & \textbf{0.104} & \textbf{41.2} & \textbf{0.0230} / \textbf{0.176} \\
\bottomrule
\end{tabular}
\end{table}

\subsection{Cascading Off-the-Shelf Tools}
\label{app:cascade}

\begin{table}[!t]
\centering
\caption{\textbf{FAIR compared with a two-stage cascade.}}
\label{tab:cascade}
\scriptsize
\renewcommand{\arraystretch}{0.9}
\begin{tabular}{lcccccc}
\toprule
& PSNR$\uparrow$ & SSIM$\uparrow$ & LPIPS\textsubscript{out}$\downarrow$ & DISTS$\downarrow$
& LPIPS\textsubscript{in}$\downarrow$ & MUSIQ$\uparrow$ \\
\midrule
HYPIR $\rightarrow$ FLUX.2-klein-9B & \textcolor{blue}{17.28} & \textcolor{blue}{0.478} & \textcolor{blue}{0.322} & \textcolor{blue}{0.162} & \textcolor{blue}{0.0482} & \textcolor{blue}{70.83} \\
\textsc{FAIR} & \textcolor{red}{20.93} & \textcolor{red}{0.559} & \textcolor{red}{0.242} & \textcolor{red}{0.139} & \textcolor{red}{0.0330} & \textcolor{red}{71.37} \\
\bottomrule
\end{tabular}
\end{table}

We compare FAIR with a cascade of HYPIR followed by unmodified
FLUX.2-klein-9B on the same inputs. In Fig.~\ref{fig:cascade}, the first
model's output has lost cues needed for faithful restoration; the second
model then copies the reference plate, producing plausible but incorrect
content. FAIR leads the cascade on every metric in Table~\ref{tab:cascade},
by $3.6$\,dB PSNR and $0.080$ LPIPS\textsubscript{out}, and on $115$ and
$113$ images respectively ($p<10^{-19}$, Wilcoxon signed-rank). This
comparison supports supplying region and reference guidance together
during restoration.

\section{Blinded Subjective Study: Protocol and Full Results}
\label{app:userstudy}

\textbf{Platform and anonymization.} The study ran on a web platform.
For each trial the nine methods are rendered as an unlabeled grid;
the slot-to-method assignment is re-randomized \emph{per viewer and
per trial}, and the true identities are stored only server-side, so
an observer cannot infer a method from its position. Observers were
not told which method was ours, nor how many methods were being
compared against a proposed one.

\textbf{Task and instructions.} For every trial the observer selects
a best and a second-best result under each of three independently
presented criteria: (i) overall perceptual quality---naturalness,
detail quality, and absence of over-generation; (ii) target-region
correction---whether the indicated local failure was fixed and
whether the result is consistent with the provided external
condition; and (iii) non-target preservation---whether content
outside the indicated region remained perceptually stable. The three
criteria are answered separately, so a method can win one and lose
another.

\textbf{Participants and coverage.} Twenty participants took part,
each evaluating $10$ cases sampled at random from our test set, for
$200$ evaluations per dimension and $600$ judgments in total.

\textbf{Statistics.} Top-2 rate is the fraction of evaluations in
which a method is chosen best or second-best; intervals are
two-sided $95\%$ Wilson score intervals ($z = 1.96$). Pairwise win
rates are derived from the same rankings: for a given trial and
dimension, a method is ranked $1$, $2$, or ``unranked'', and FAIR
beats a baseline when it holds the strictly better rank. Trials in
which neither method is ranked carry no ordering information and are
excluded from that pair.

\begin{table}[h]
\centering
\caption{\textbf{\boldmath Pairwise win rates of FAIR against each baseline,
derived from the blinded top-2 rankings, with two-sided $95\%$
Wilson intervals.} FAIR wins against every baseline on all three
dimensions, and every interval lower bound is at least $73.9\%$.}
\label{tab:userstudy_pairwise}
\scriptsize
\renewcommand{\arraystretch}{0.9}
\setlength{\tabcolsep}{4pt}
\begin{tabular}{lccc}
\toprule
Baseline & Overall quality & Target correction & Non-target preservation \\
\midrule
InstructRestore & 99.4\% \scriptsize(96.5--99.9) & 99.4\% \scriptsize(96.5--99.9) & 99.4\% \scriptsize(96.5--99.9) \\
SeeSR           & 95.2\% \scriptsize(90.7--97.5) & 96.3\% \scriptsize(92.1--98.3) & 95.1\% \scriptsize(90.6--97.5) \\
AdaRefSR        & 94.6\% \scriptsize(90.0--97.1) & 93.2\% \scriptsize(88.3--96.2) & 93.4\% \scriptsize(88.5--96.3) \\
ReFIR+SUPIR     & 93.4\% \scriptsize(88.6--96.3) & 91.6\% \scriptsize(86.4--94.9) & 92.1\% \scriptsize(87.0--95.3) \\
ReFIR+SeeSR     & 91.1\% \scriptsize(85.8--94.5) & 91.1\% \scriptsize(85.8--94.5) & 91.1\% \scriptsize(85.8--94.5) \\
CODSR           & 89.6\% \scriptsize(84.2--93.3) & 92.8\% \scriptsize(87.9--95.8) & 91.8\% \scriptsize(86.7--95.1) \\
HYPIR           & 84.5\% \scriptsize(78.6--89.1) & 85.1\% \scriptsize(79.1--89.7) & 83.2\% \scriptsize(77.1--88.0) \\
SUPIR           & 80.5\% \scriptsize(73.9--85.7) & 86.2\% \scriptsize(80.3--90.6) & 88.4\% \scriptsize(82.7--92.3) \\
\bottomrule
\end{tabular}
\end{table}

\section{Robustness of the Automatic Condition Pipeline}
\label{app:robustness}

The automatic pipeline uses Qwen3-VL, SAM3, and DINOv3/CLIP
to prepare restoration conditions from the low-quality input. We compare
it with a GT-assisted condition oracle, smaller condition models,
mismatched references, and repeated runs to assess the stability and
contribution of these components.

\textbf{Condition prediction accuracy.}
On a held-out split, the fine-tuned Qwen3-VL-8B predicts the failure type
from $\mathbf{x}_\text{LQ}$ alone with $84.03\%$ accuracy, and $90.66\%$ of
the resulting masks reach IoU $\geq 0.5$ against the annotated masks.

\textbf{LQ-only inference versus a GT-assisted oracle.}
In the oracle setting the \emph{only} change is that Qwen receives
$(\mathbf{x}_\text{LQ}, \mathbf{x}_\text{GT})$ instead of
$\mathbf{x}_\text{LQ}$; SAM3 still turns the predicted region phrase
into a mask, and reference acquisition and restoration are identical.
Table~\ref{tab:lq_vs_oracle} shows that the fully automatic LQ-only
pipeline is close to this oracle on every metric, confirming
effective GT-free operation.

\begin{table}[!ht]
\centering
\caption{\textbf{LQ-only inference versus the GT-assisted condition
oracle.} The oracle differs only in that Qwen additionally sees
$\mathbf{x}_\text{GT}$ when predicting the failure region and type;
differences are within $0.35\%$ on all
ten metrics, and within $0.12\%$ on the no-reference and
reference-similarity scores.}
\label{tab:lq_vs_oracle}
\scriptsize
\renewcommand{\arraystretch}{0.9}
\setlength{\tabcolsep}{4pt}
\begin{tabular}{lccccc}
\toprule
& PSNR$\uparrow$ & SSIM$\uparrow$ & LPIPS\textsubscript{out}$\downarrow$ & FID$\downarrow$
& LPIPS\textsubscript{in}$\downarrow$  \\
\midrule
LQ-only (ours)      & 20.91 & 0.5582 & 0.2429 & 63.28 & 0.0319 \\
GT-assisted oracle  & 20.91 & 0.5584 & 0.2432 & 63.22 & 0.0320 \\
\midrule
& MUSIQ$\uparrow$ & CLIP-IQA$\uparrow$ & Q-Align$\uparrow$
& DINO-Sim-Ref$\uparrow$ & CLIP-Sim-Ref$\uparrow$ \\
\midrule
LQ-only (ours)      & 71.47 & 0.6728 & 4.514 & 0.4677 & 0.5981 \\
GT-assisted oracle  & 71.50 & 0.6736 & 4.517 & 0.4682 & 0.5976 \\
\bottomrule
\end{tabular}
\end{table}

\textbf{Smaller condition models.}
Replacing Qwen3-VL-8B with Qwen3-VL-4B, and SAM3 with
SAM2.1-Hiera-Tiny ($\approx 1/15$ the parameters), leaves restoration
quality essentially unchanged (Table~\ref{tab:tool_robustness}); the
smaller segmenter still reproduces the same regions (median mask IoU
$0.873$, $93.3\%$ of regions at IoU $\geq 0.5$). The gains therefore
do not come from the scale of the external tools.

\begin{table}[!ht]
\centering
\caption{\textbf{Robustness to smaller condition models.} All other
settings fixed. Replacing the $8$B VLM with a $4$B one, or SAM3 with
SAM2.1-Hiera-Tiny ($\approx 1/15$ the parameters), leaves restoration
quality essentially unchanged.}
\label{tab:tool_robustness}
\scriptsize
\renewcommand{\arraystretch}{0.9}
\setlength{\tabcolsep}{3.5pt}
\begin{tabular}{lcccccccc}
\toprule
Condition models & PSNR$\uparrow$ & SSIM$\uparrow$ & LPIPS\textsubscript{out}$\downarrow$
& FID$\downarrow$ & LPIPS\textsubscript{in}$\downarrow$
& MUSIQ$\uparrow$ & CLIP-IQA$\uparrow$
& Q-Align$\uparrow$ \\
\midrule
Qwen3-VL-8B + SAM3 (default) & 20.91 & 0.5582 & 0.2429 & 63.28 & 0.0319 & 71.47 & 0.6728 & 4.514 \\
Qwen3-VL-4B + SAM3           & 20.91 & 0.5581 & 0.2431 & 63.26 & 0.0319 & 71.49 & 0.6728 & 4.514 \\
Qwen3-VL-8B + SAM2.1-Tiny    & 20.92 & 0.5590 & 0.2422 & 62.11 & 0.0313 & 71.48 & 0.6726 & 4.518 \\
\bottomrule
\end{tabular}
\end{table}

\begingroup\setlength{\intextsep}{2pt}\setlength{\textfloatsep}{4pt}\setlength{\floatsep}{4pt}\setlength{\abovecaptionskip}{2pt}\setlength{\belowcaptionskip}{0pt}\textbf{Mismatched references.}
Shuffling references within each failure category (Table~\ref{tab:ref_shuffle}) changes PSNR, SSIM, MUSIQ, CLIP-IQA, and Q-Align by at most $0.91\%$, but affects FID, LPIPS, and reference similarity more, confirming that FAIR uses the reference.

\makeatletter\setlength{\@fptop}{0pt}\setlength{\@fpsep}{14pt}\setlength{\@fpbot}{0pt plus 1fil}\makeatother

\begin{table}[!ht]
\centering
\caption{\textbf{Robustness to mismatched references}, using the benchmark's
region annotations as in Table~\ref{tab:ablation}. References are
shuffled within each failure category. PSNR, SSIM, MUSIQ, CLIP-IQA, and
Q-Align change by at most $0.91\%$; FID, LPIPS, and reference-similarity
scores show larger changes.}
\label{tab:ref_shuffle}
\scriptsize
\renewcommand{\arraystretch}{0.9}
\setlength{\tabcolsep}{5pt}
\begin{tabular}{lcc|lcc}
\toprule
Metric & Default & Shuffled & Metric & Default & Shuffled \\
\midrule
PSNR$\uparrow$     & 20.91 & 20.80 \;\scriptsize($-0.52\%$) & LPIPS\textsubscript{out}$\downarrow$ & 0.2432 & 0.2490 \;\scriptsize($+2.38\%$) \\
SSIM$\uparrow$     & 0.5584 & 0.5533 \;\scriptsize($-0.91\%$) & FID$\downarrow$          & 63.22 & 68.67 \;\scriptsize($+8.63\%$) \\
MUSIQ$\uparrow$    & 71.50 & 71.57 \;\scriptsize($+0.10\%$) & LPIPS\textsubscript{in}$\downarrow$ & 0.0320 & 0.0353 \;\scriptsize($+10.32\%$) \\
CLIP-IQA$\uparrow$ & 0.6736 & 0.6749 \;\scriptsize($+0.18\%$) & DINO-Sim-Ref$\uparrow$   & 0.4682 & 0.4513 \;\scriptsize($-3.61\%$) \\
Q-Align$\uparrow$  & 4.5168 & 4.5040 \;\scriptsize($-0.28\%$) & CLIP-Sim-Ref$\uparrow$   & 0.5976 & 0.5884 \;\scriptsize($-1.54\%$) \\
\bottomrule
\end{tabular}
\end{table}

\textbf{Irrelevant and wrong references.} Table~\ref{tab:wrong_ref} gives all metrics for the mismatched-reference settings of Sec.~\ref{sec:analysis}.

\begin{table}[!ht]
\centering
\caption{\textbf{Irrelevant and wrong references.} DINO and CLIP similarities are computed between the output's target-region crop and the supplied $v$; content matches use the judge of Appendix~\ref{app:semantic}.}
\label{tab:wrong_ref}
\scriptsize
\renewcommand{\arraystretch}{0.9}
\setlength{\tabcolsep}{2.5pt}
\begin{tabular}{lcccccccccc}
\toprule
Visual guidance & PSNR$\uparrow$ & FID$\downarrow$ & LPIPS$\downarrow$ & LPIPS\textsubscript{in}$\downarrow$ & LPIPS\textsubscript{out}$\downarrow$ & DISTS$\downarrow$ & MUSIQ$\uparrow$ & DINO sim. & CLIP sim. & \makecell{Content\\matches$\uparrow$} \\
\midrule
Table~\ref{tab:quant_comparison_synthetic} reference & 20.91 & 63.28 & 0.2703 & 0.0319 & 0.2429 & 0.1387 & 71.47 & 0.650 & 0.680 & 107/120 \\
Another failure category & 20.82 & 68.06 & 0.2788 & 0.0349 & 0.2488 & 0.1455 & 71.41 & 0.020 & 0.266 & 86/120 \\
Unrelated crop & 20.82 & 68.37 & 0.2791 & 0.0351 & 0.2490 & 0.1462 & 71.45 & 0.030 & 0.351 & 92/120 \\
Omitted & 20.84 & 67.76 & 0.2782 & 0.0350 & 0.2484 & 0.1458 & 71.16 & -- & -- & 93/120 \\
\bottomrule
\end{tabular}
\end{table}

\textbf{Visual-guidance abstention.} At inference, the retrieval of Appendix~\ref{app:reference_retrieval} omits the reference when no candidate reaches visual similarity $0.50$ and passes the Qwen3-VL semantic check. FAIR then receives the black-image condition that it sees as dropout in training (Sec.~\ref{sec:curriculum}) and restores the region from $R$ and $t$. In this no-reference setting, FAIR obtains FID $67.76$ and LPIPS\textsubscript{in}/LPIPS\textsubscript{out} $0.0350/0.2484$ on all $120$ images, ahead of the best baseline values ($81.93$, $0.0367/0.272$).

\textbf{Region control.} Region control specifies which object receives guidance. When similar objects fit the same description, the instruction alone cannot distinguish them. In strict-region mode, FAIR uses reference-guided restoration inside $R$ and reference-free restoration outside, so only the selected boy receives the plaid pattern (Fig.~\ref{fig:multiregion}, bottom). Image- and region-level averages change little without $R$ because LPIPS spatially averages feature differences, diluting changes confined to one object. Table~\ref{tab:mask_ablation} reports results with and without $R$.

\begin{table}[!ht]
\centering
\caption{\textbf{Results with and without the region map $R$.} Both columns use the same model weights and degraded inputs; \emph{Without $R$} removes the region map.}
\label{tab:mask_ablation}
\scriptsize
\renewcommand{\arraystretch}{0.9}
\begin{tabular}{lcc}
\toprule
Metric & With $R$ & Without $R$ \\
\midrule
LPIPS\textsubscript{in}$\downarrow$ & 0.0316 & 0.0319 \\
PSNR\textsubscript{out} (dB)$\uparrow$ & 21.20 & 21.18 \\
MUSIQ$\uparrow$ & 71.54 & 71.50 \\
DINO sim.\ to $v$$\uparrow$ & 0.659 & 0.655 \\
CLIP sim.\ to $v$$\uparrow$ & 0.684 & 0.683 \\
LPIPS\textsubscript{out}$\downarrow$ & 0.2431 & 0.2430 \\
\bottomrule
\end{tabular}
\end{table}

\subsection{Run-to-Run Variation}
\label{app:errorbars}

Table~\ref{tab:errorbars} reports mean $\pm$ standard deviation over
five inference runs with different random seeds on the
synthetic benchmark. Relative variation stays below $0.6\%$ on every
metric, so the margins reported in
Table~\ref{tab:quant_comparison_synthetic} are not attributable to
run-to-run noise. These runs vary sampling on a fixed test set. Because an FID estimated from $120$ images also depends on which images are included, we corroborate it with per-region LPIPS, the blinded study (Sec.~\ref{sec:userstudy}), and the $287$ images of WR-SR and the real-world set (Appendix~\ref{app:experimental_results}); on WR-SR, FAIR also attains the lowest FID and LPIPS.

\begin{table}[!ht]
\centering
\caption{\textbf{\boldmath Mean $\pm$ standard deviation over five runs.}
Relative standard deviation in parentheses.}
\label{tab:errorbars}
\scriptsize
\renewcommand{\arraystretch}{0.9}
\setlength{\tabcolsep}{6pt}
\begin{tabular}{lc|lc}
\toprule
Metric & Mean $\pm$ Std. & Metric & Mean $\pm$ Std. \\
\midrule
PSNR         & $20.93 \pm 0.009$ \scriptsize(0.04\%)  & Q-Align       & $4.508 \pm 0.0056$ \scriptsize(0.12\%) \\
SSIM         & $0.5586 \pm 0.0002$ \scriptsize(0.03\%) & MUSIQ         & $71.45 \pm 0.039$ \scriptsize(0.05\%) \\
LPIPS\textsubscript{out} & $0.2432 \pm 0.0003$ \scriptsize(0.10\%) & CLIP-IQA      & $0.6713 \pm 0.0021$ \scriptsize(0.31\%) \\
FID          & $62.84 \pm 0.33$ \scriptsize(0.52\%)    & DINO-Sim-Ref  & $0.4690 \pm 0.0008$ \scriptsize(0.16\%) \\
LPIPS\textsubscript{in} & $0.0319 \pm 0.00005$ \scriptsize(0.15\%) & CLIP-Sim-Ref & $0.5989 \pm 0.0007$ \scriptsize(0.12\%) \\

\bottomrule
\end{tabular}
\end{table}

\endgroup
\clearpage

\section{Additional Qualitative Comparisons}
\label{app:qualitative_results}

\begin{figure*}[p]
  \centering
  \vspace{-20pt}
  \includegraphics[width=\linewidth]{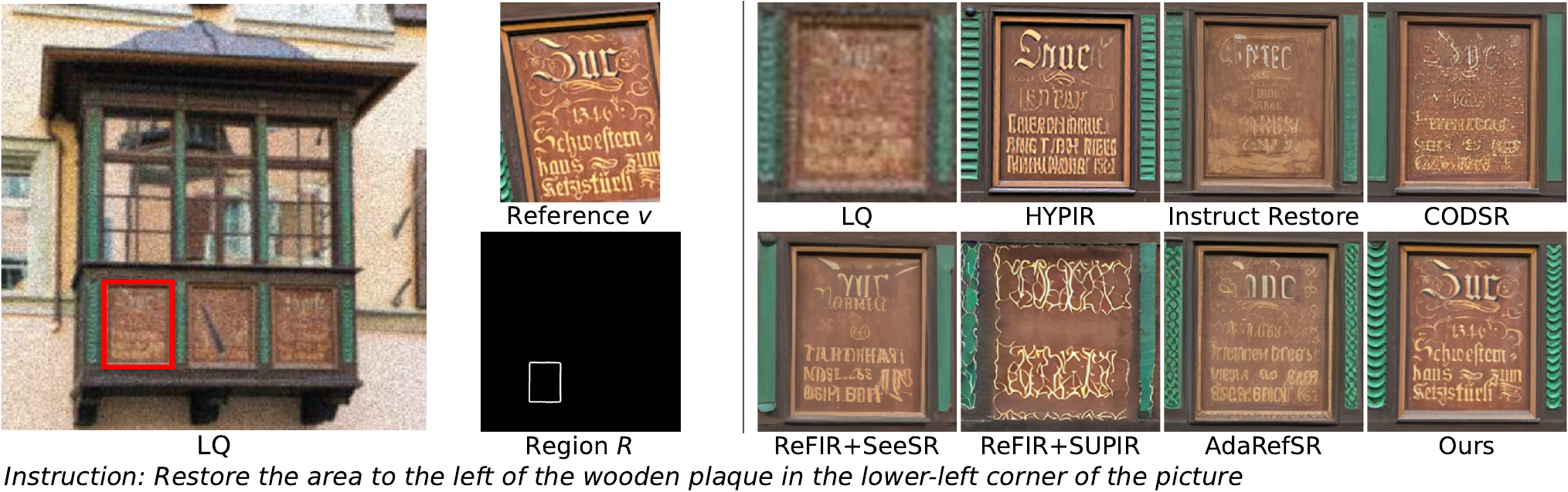}\\
  \includegraphics[width=\linewidth]{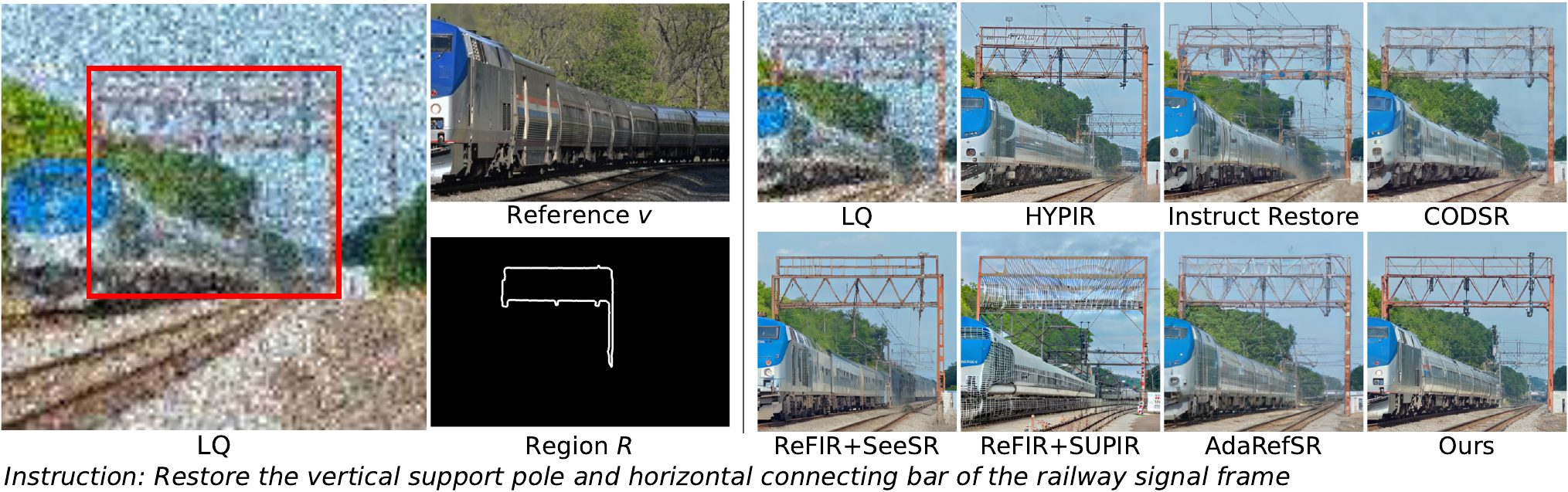}\\
  \includegraphics[width=\linewidth]{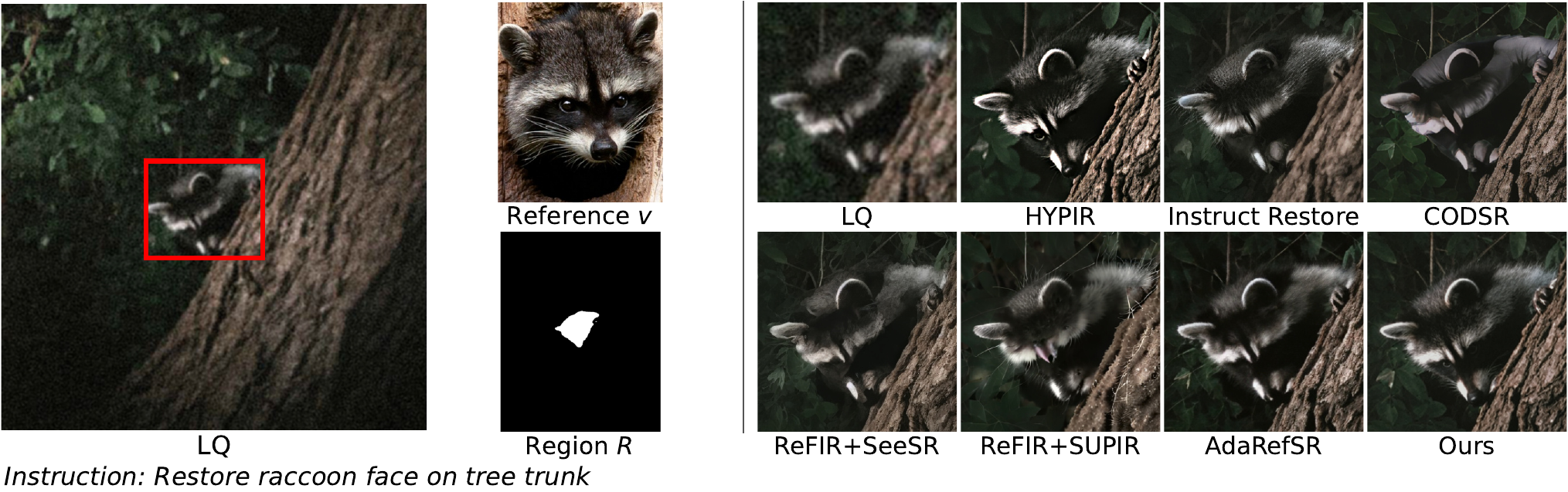}\\
  \includegraphics[width=\linewidth]{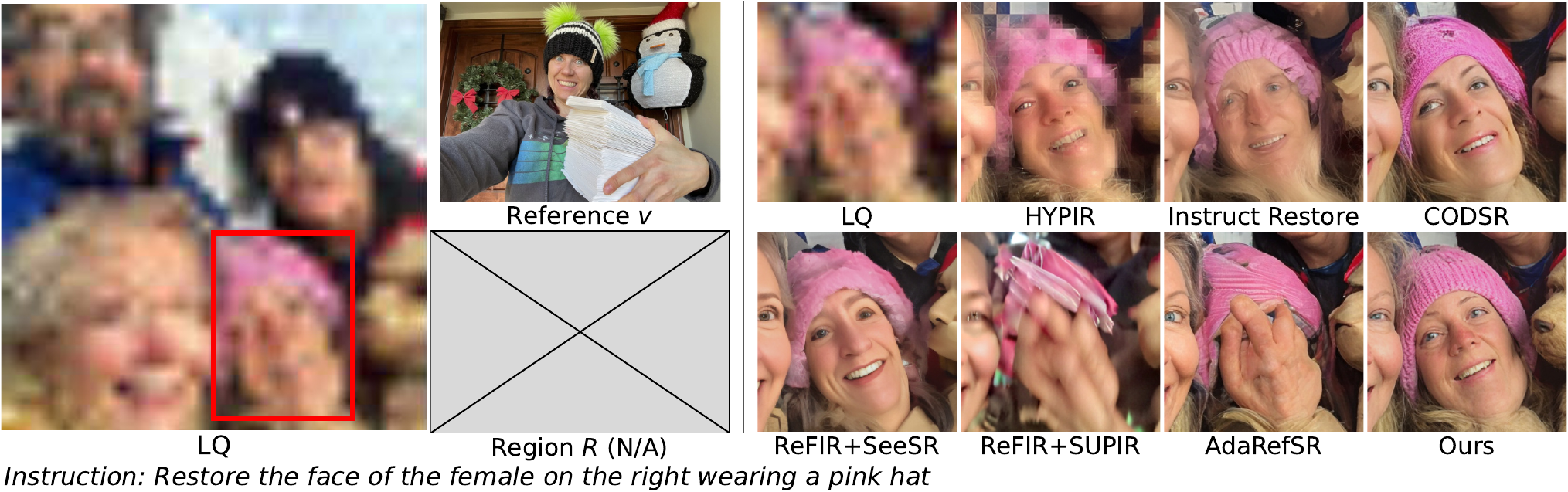}\\
  \includegraphics[width=\linewidth]{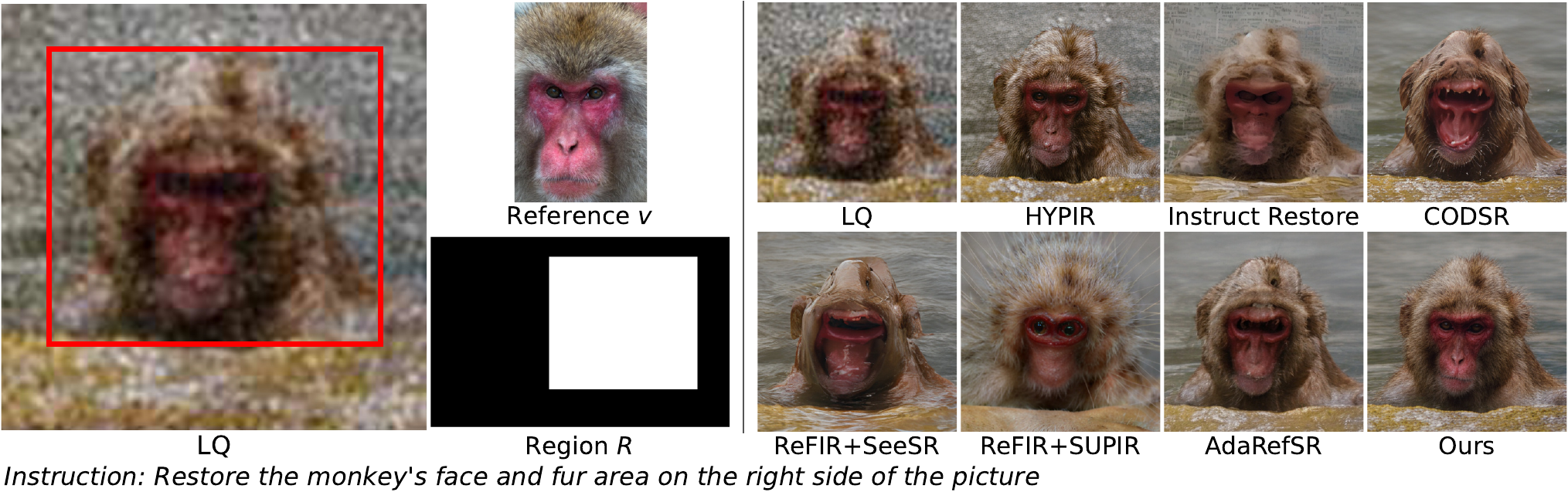}\\
  \vspace{-5pt}
  \caption{\textbf{Additional qualitative comparisons on the proposed test set (1/2).} Each case shows the LQ input with the target region (red box), the visual guidance $v$, the region control $R$, and the textual instruction $t$, followed by the outputs of the compared methods.}
  \label{fig:comp_appx1}
  \vspace{-10pt}
\end{figure*}

\begin{figure*}[p]
  \centering
  \vspace{-20pt}
  \includegraphics[width=\linewidth]{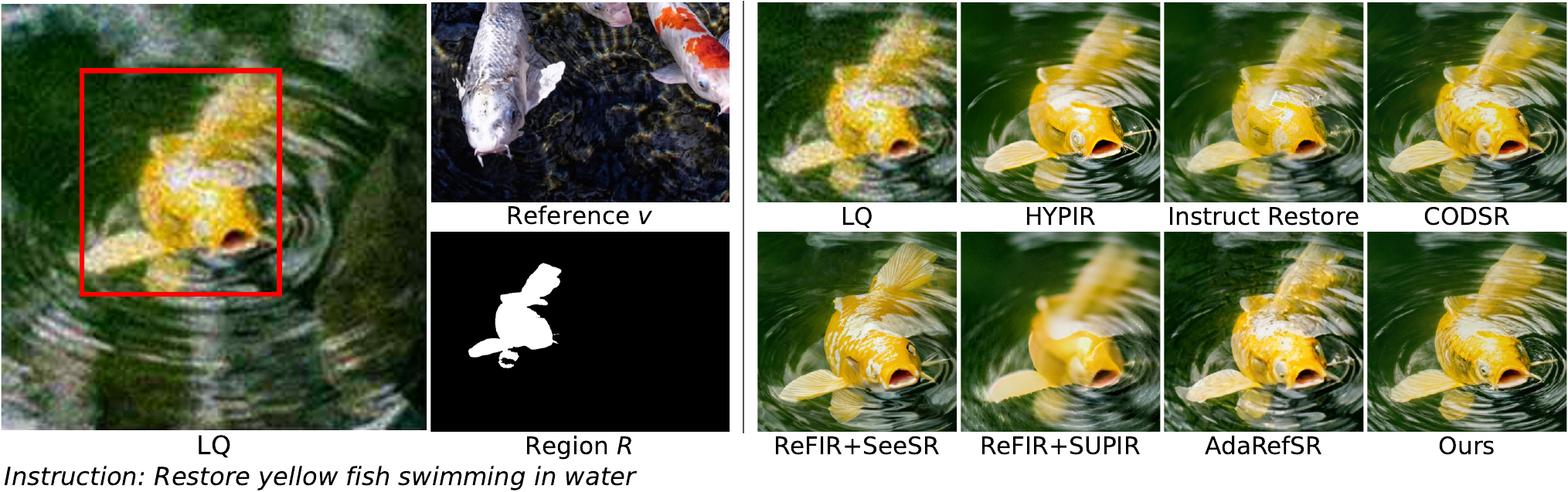}\\
  \includegraphics[width=\linewidth]{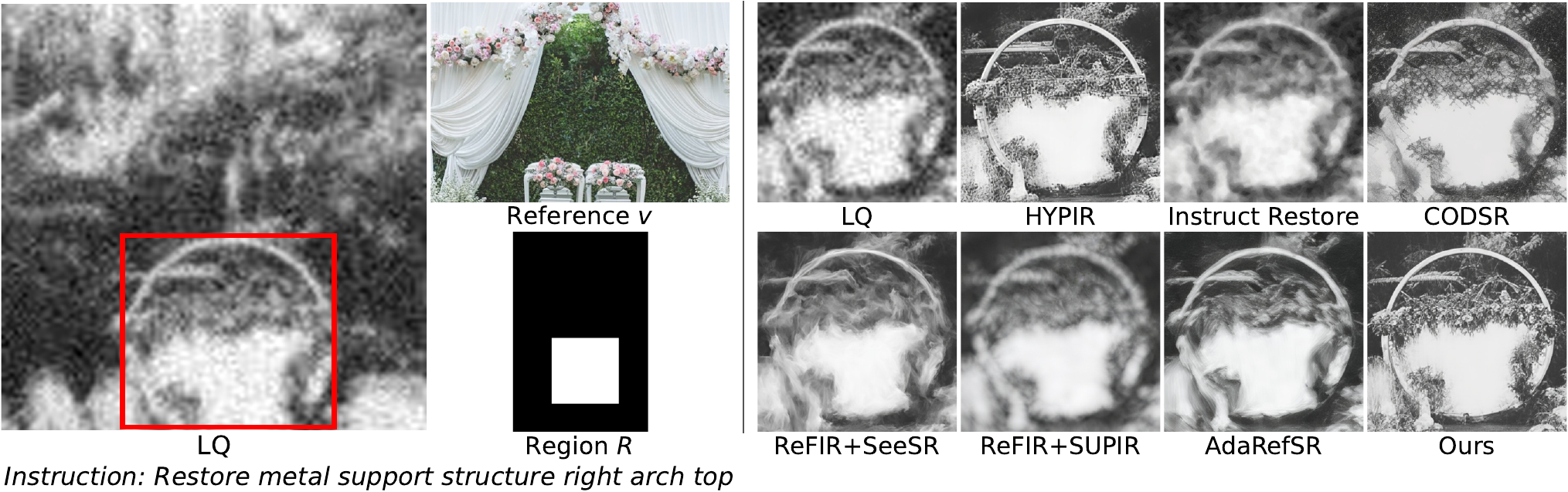}\\
  \includegraphics[width=\linewidth]{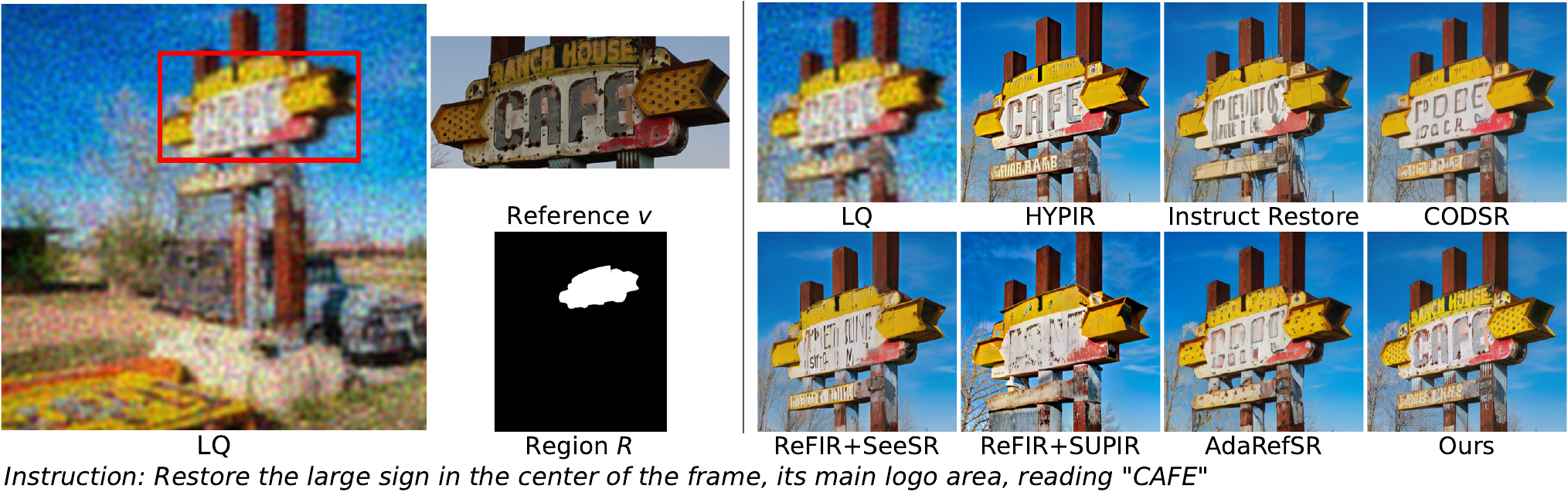}\\
  \includegraphics[width=\linewidth]{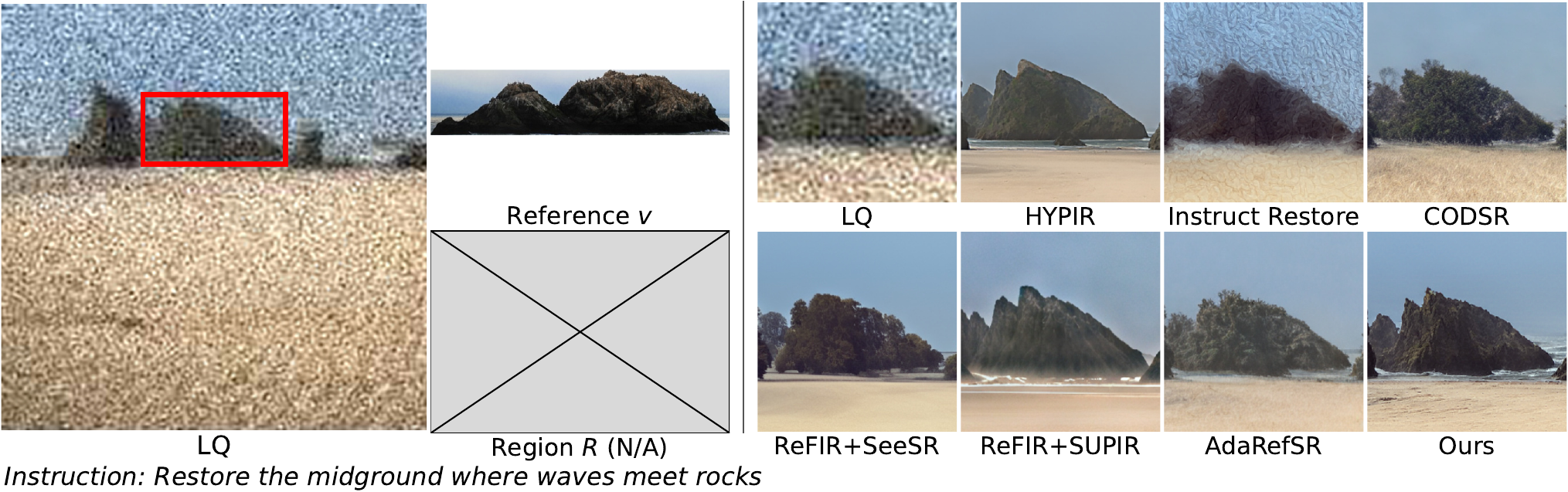}\\
  \includegraphics[width=\linewidth]{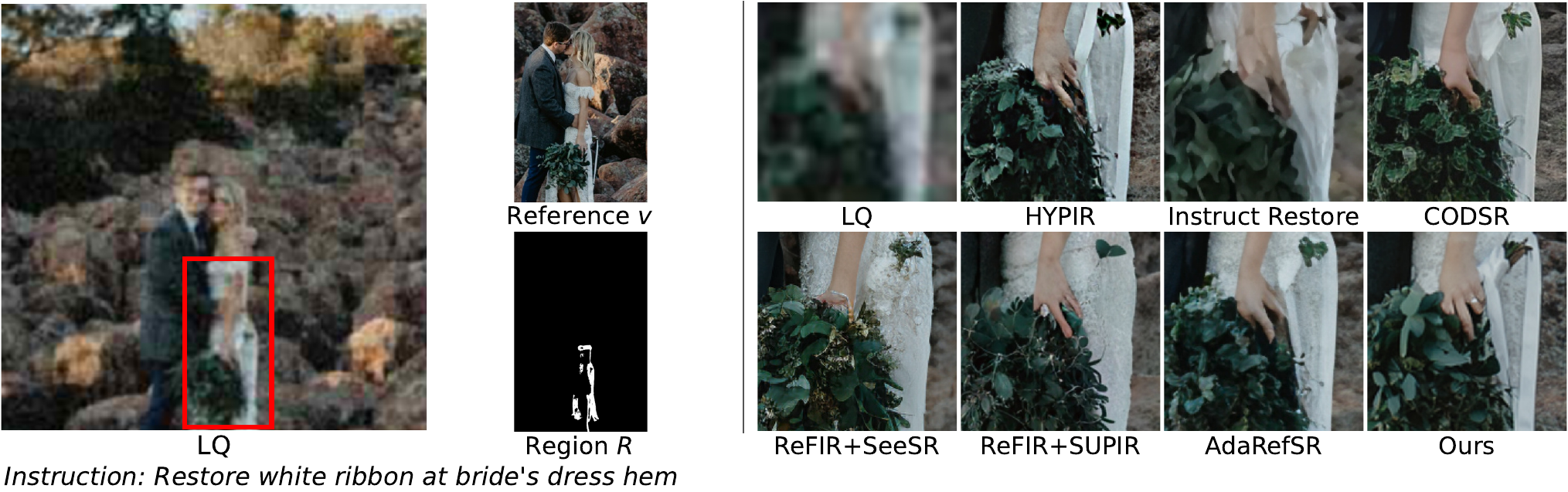}\\
  \vspace{-5pt}
  \caption{\textbf{Additional qualitative comparisons on the proposed test set (2/2).} Each case shows the LQ input with the target region (red box), the visual guidance $v$, the region control $R$, and the textual instruction $t$, followed by the outputs of the compared methods.}
  \label{fig:comp_appx2}
  \vspace{-10pt}
\end{figure*}

\begin{figure*}[p]
  \centering
  \vspace{-20pt}
  \includegraphics[width=\linewidth]{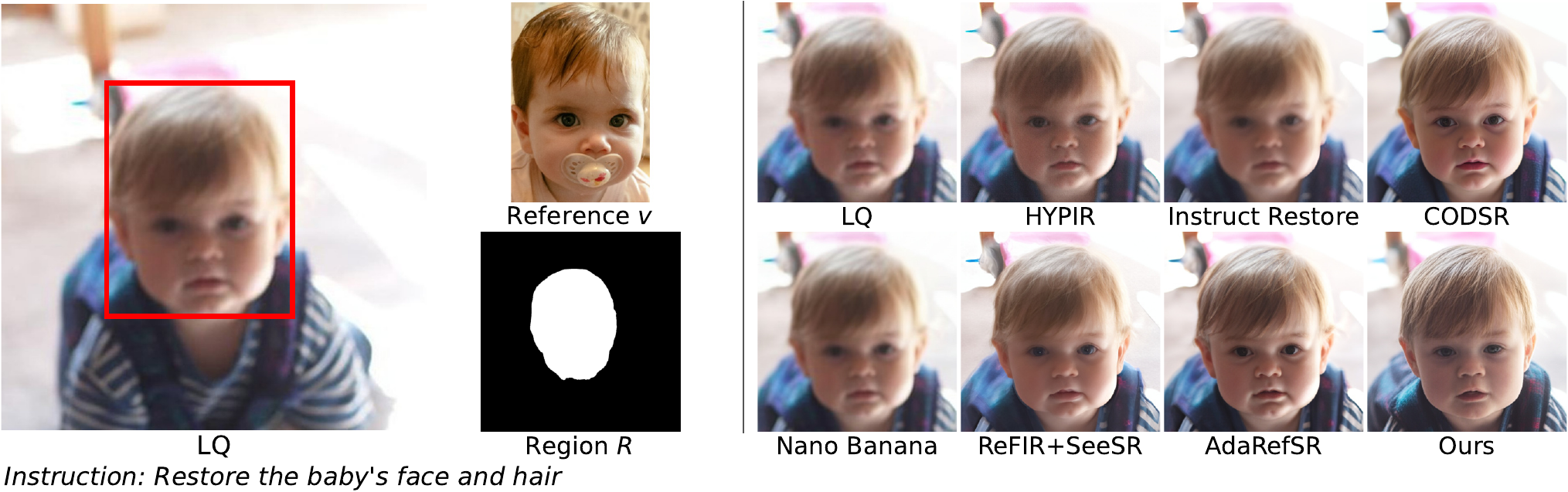}\\
  \includegraphics[width=\linewidth]{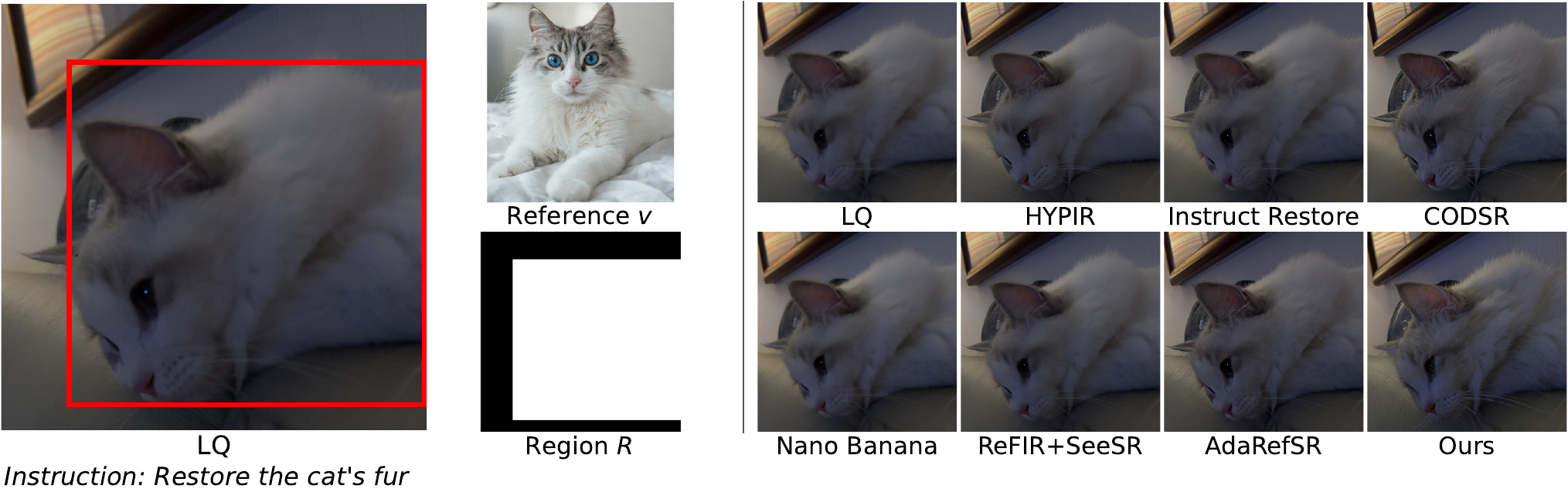}\\
  \includegraphics[width=\linewidth]{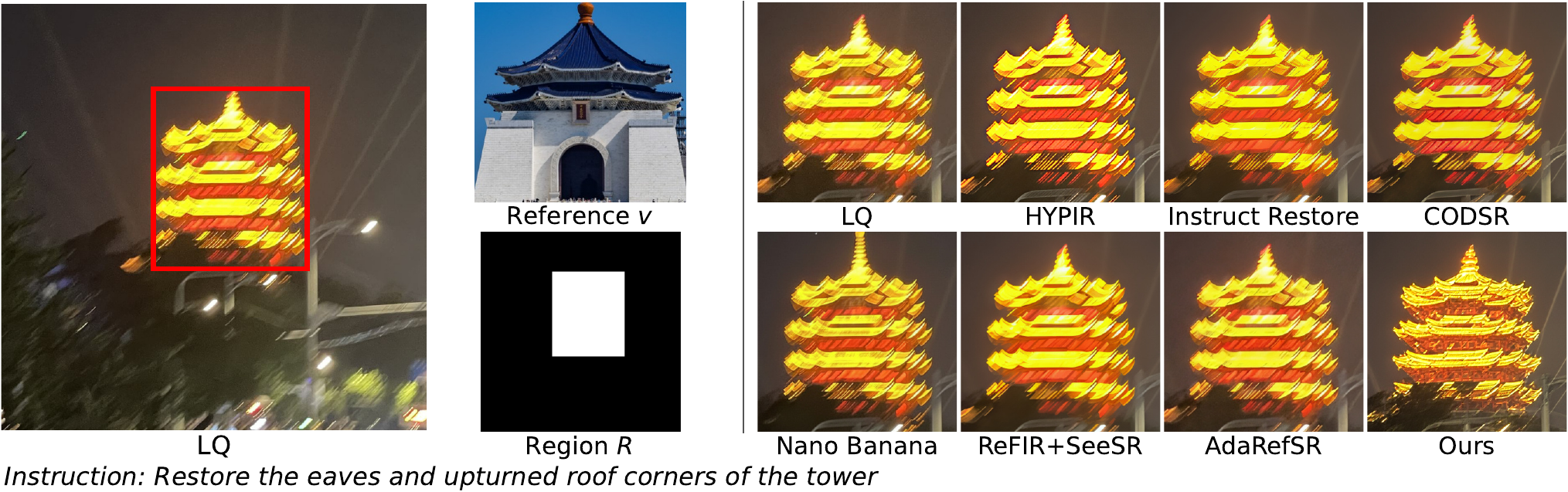}\\
  \includegraphics[width=\linewidth]{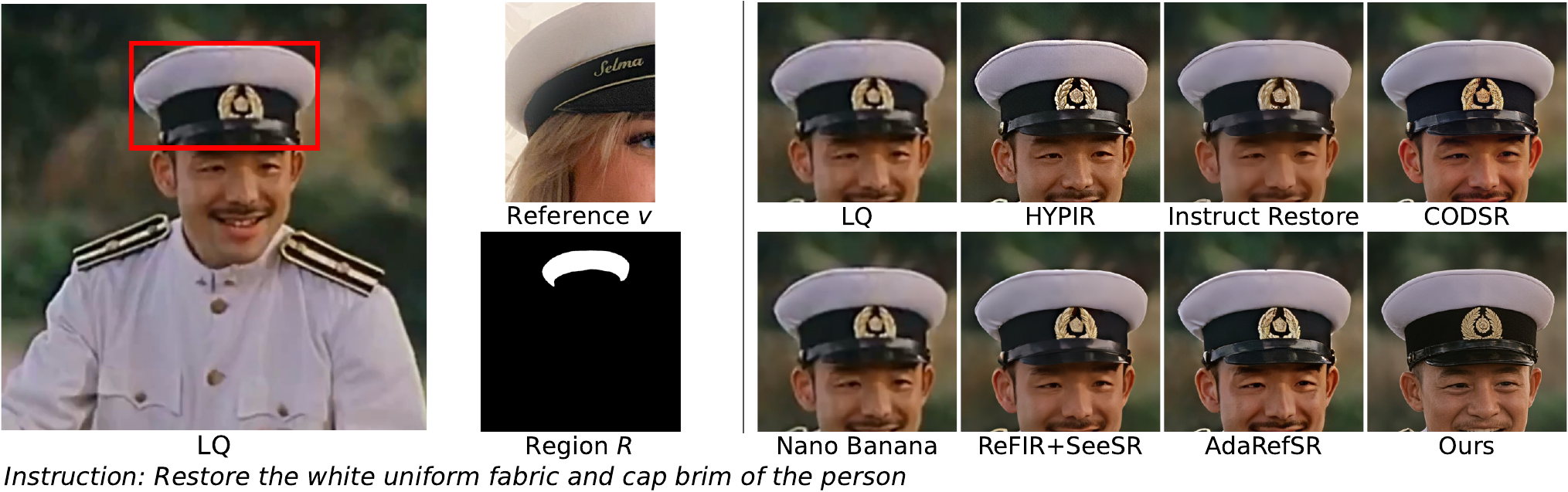}\\
  \includegraphics[width=\linewidth]{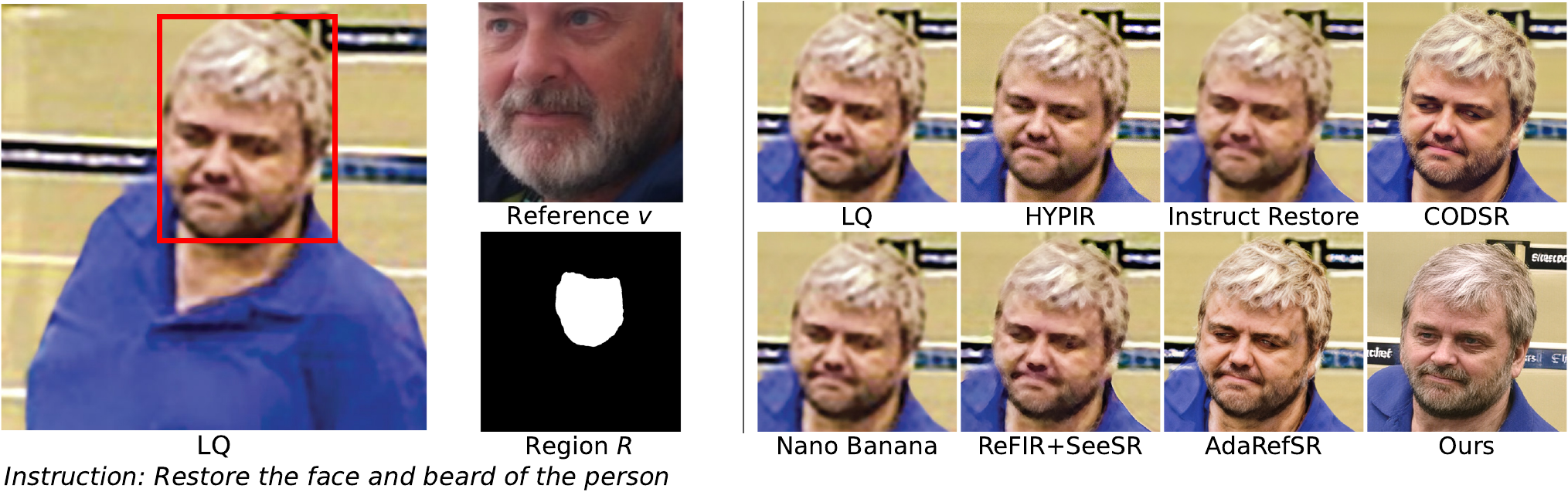}\\
  \vspace{-5pt}
  \caption{\textbf{Additional qualitative comparisons on the real-world test set.} Each case shows the LQ input with the target region (red box), the visual guidance $v$, the region control $R$, and the textual instruction $t$, followed by the outputs of the compared methods.}
  \label{fig:comp_appx_ext}
  \vspace{-10pt}
\end{figure*}

\begin{figure*}[p]
  \centering
  \vspace{-20pt}
  \includegraphics[width=\linewidth]{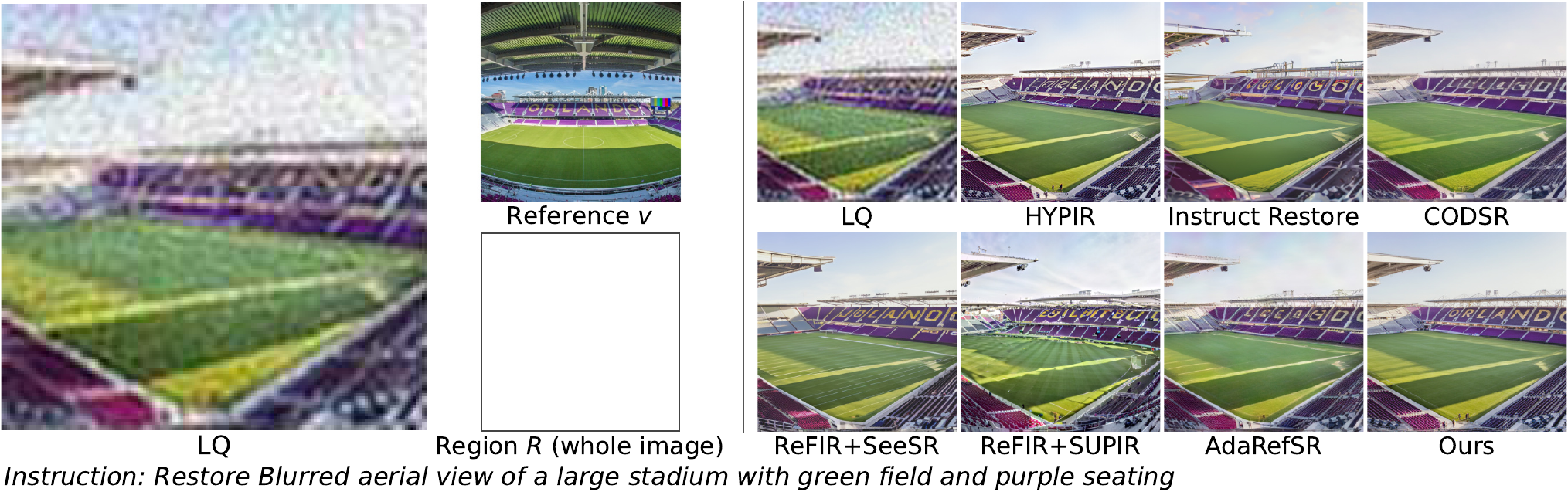}\\
  \includegraphics[width=\linewidth]{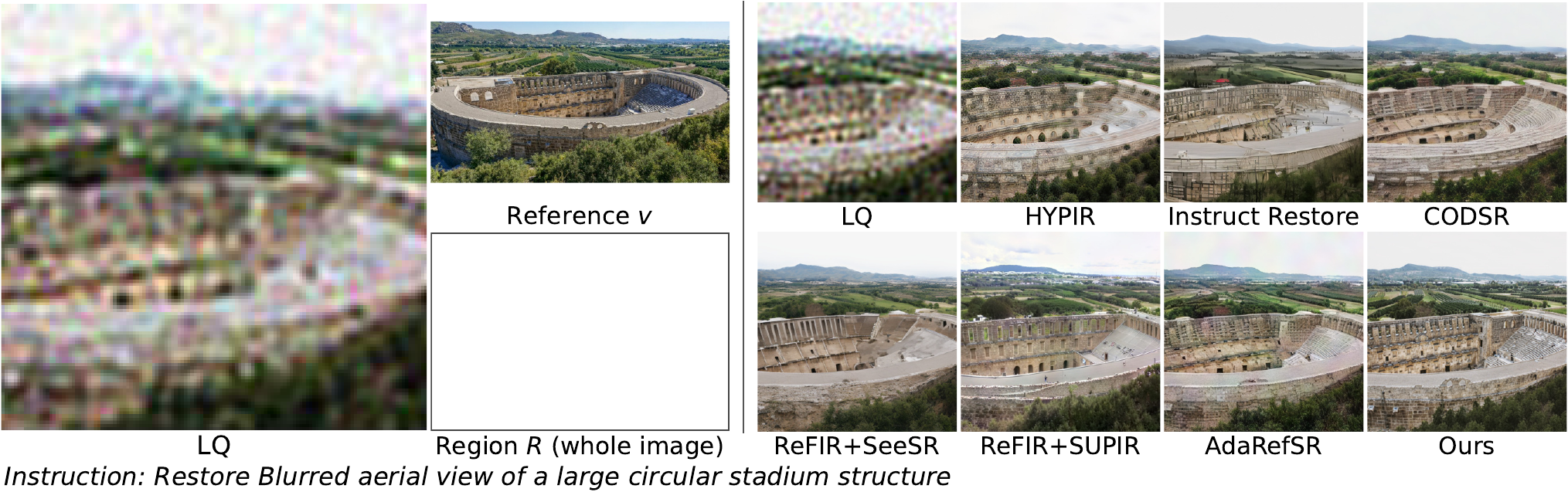}\\
  \includegraphics[width=\linewidth]{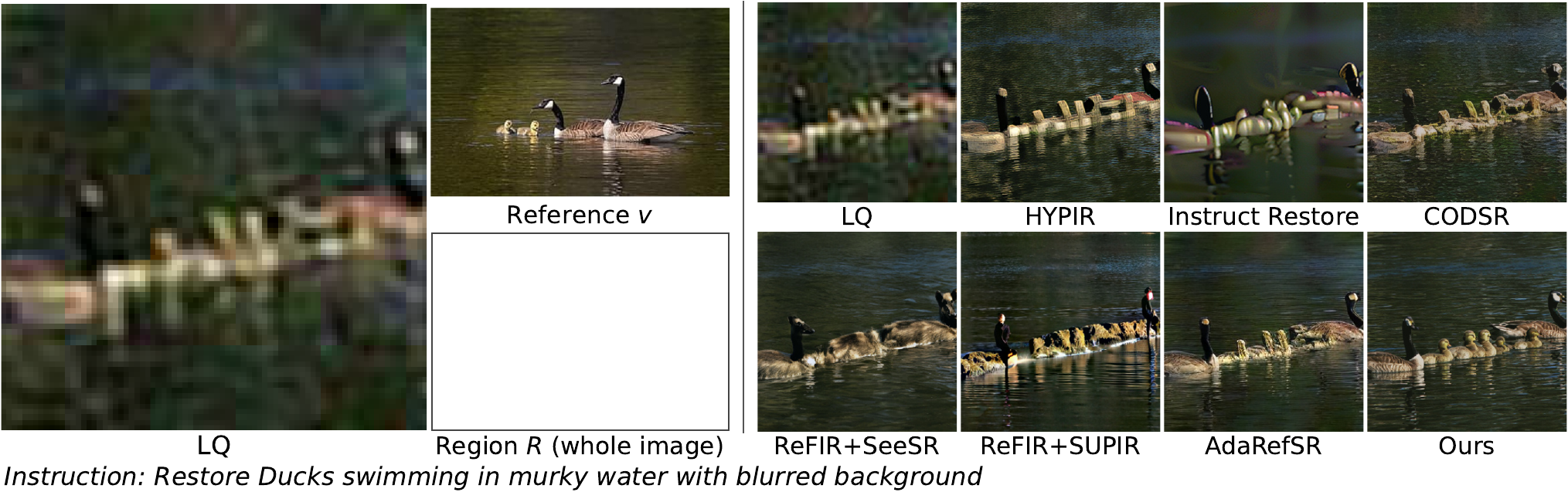}\\
  \includegraphics[width=\linewidth]{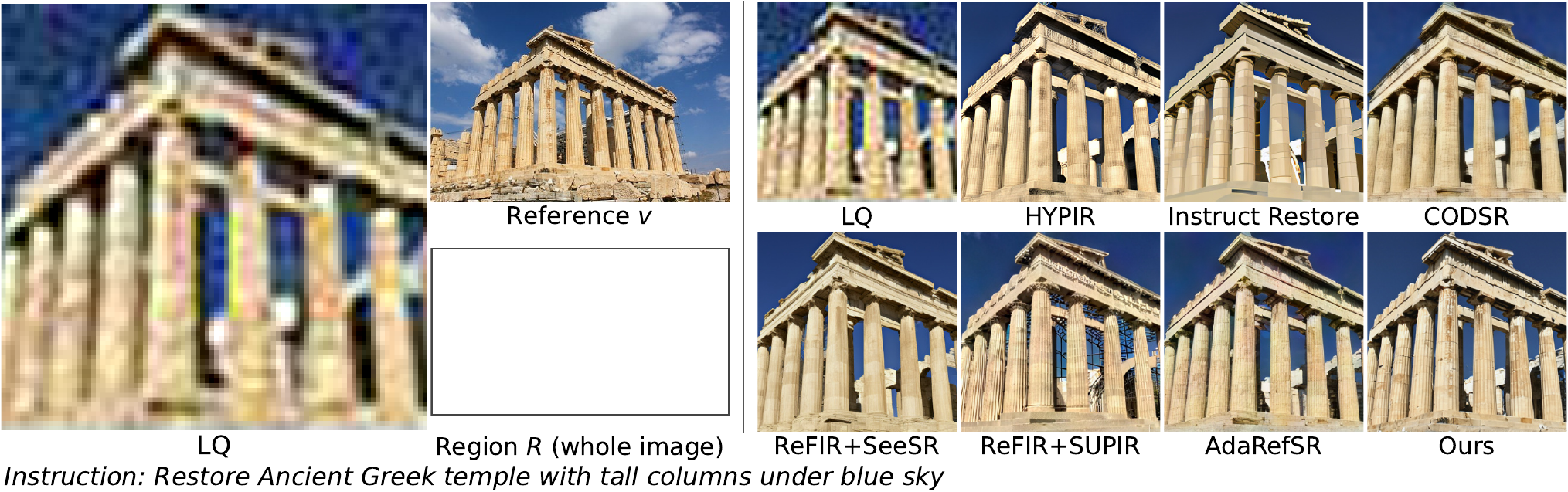}\\
  \includegraphics[width=\linewidth]{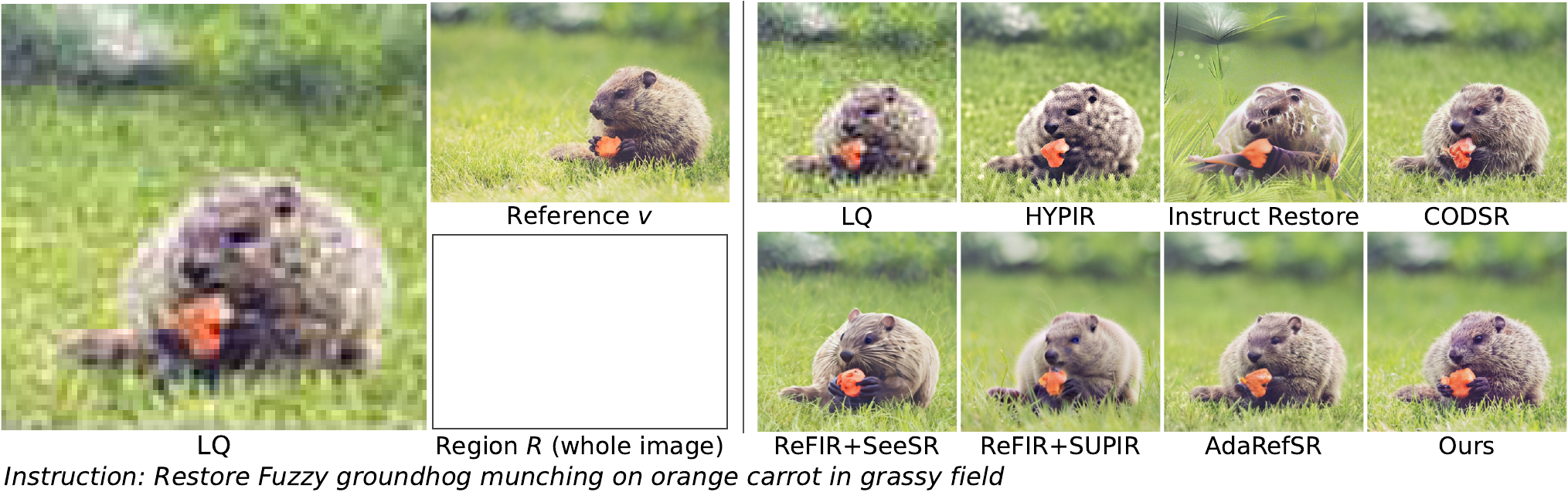}\\
  \vspace{-5pt}
  \caption{\textbf{Additional qualitative comparisons on WR-SR.} Each case shows the LQ input, the curated reference $v$, the region control $R$, which covers the whole image, and the textual instruction $t$, followed by the outputs of the compared methods.}
  \label{fig:comp_appx_wrsr}
  \vspace{-10pt}
\end{figure*}

Figures~\ref{fig:comp_appx1} and~\ref{fig:comp_appx2} show further cases from the proposed
test set, and Figures~\ref{fig:comp_appx_ext} and~\ref{fig:comp_appx_wrsr} show cases from the
real-world test set and WR-SR. Each case gives the LQ input, the visual guidance $v$, the region control $R$ and the
textual instruction $t$, followed by the outputs of the compared methods. As in
Figure~\ref{fig:comp_main}, FAIR follows the reference inside the target region while
preserving the surroundings.

The cascade comparison discussed in Appendix~\ref{app:cascade} is shown in Fig.~\ref{fig:cascade}.

\begin{figure*}[!ht]
  \centering
  \includegraphics[width=\linewidth]{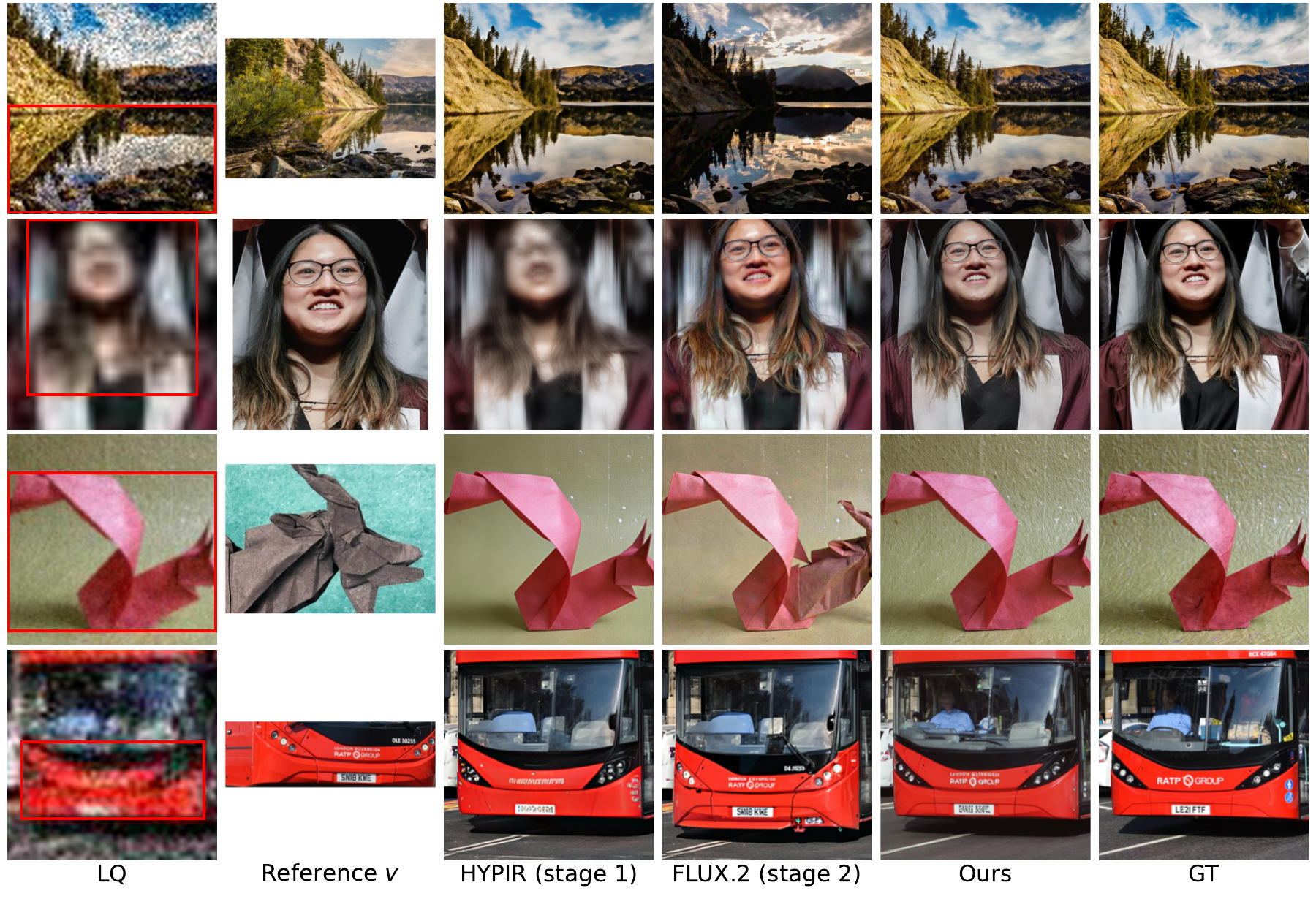}
  \vspace{-14pt}
  \caption{\textbf{FAIR compared with a two-stage cascade.} The second model
  receives the first model's output. In the last row, it copies the plate
  \texttt{SNI8 KWE} from the reference, whereas the GT reads
  \texttt{LE21 FTF}.}
  \vspace{-10pt}
  \label{fig:cascade}
\end{figure*}

\clearpage

\end{document}